\documentclass[10pt,twocolumn,letterpaper]{article}

\usepackage[pagenumbers]{iccv} 

\usepackage{dsfont}
\usepackage{multirow}
\usepackage{multicol}
\usepackage{subcaption}
\usepackage{floatrow}
\usepackage{pifont}
\usepackage{stfloats}
\usepackage{pgfplots}
\usepackage{pgfplotstable}
\pgfplotsset{compat=1.18} 
\usepackage{colortbl}

\newcommand{\keypoint}[1]{\vspace{0.1cm}\noindent\textbf{#1}\quad}
\newcommand{\cmark}{\ding{51}}%
\newcommand{\xmark}{\ding{55}}%

\definecolor{Gray}{gray}{0.9}

\newcolumntype{g}{>{\columncolor{Gray}}c}

\definecolor{iccvblue}{rgb}{0.21,0.49,0.74}
\usepackage[pagebackref,breaklinks,colorlinks,allcolors=iccvblue]{hyperref}

\def\paperID{02922} 
\def\confName{ICCV}
\def\confYear{2025}

\title{\vspace{-2mm}Doodle Your Keypoints: Sketch-Based Few-Shot Keypoint Detection\vspace{-2mm}}

\author{
Subhajit Maity\textsuperscript{1}\thanks{Work done as an intern at SketchX before joining UCF.}\hspace{.2cm}
Ayan Kumar Bhunia\textsuperscript{2}\hspace{.2cm}
Subhadeep Koley\textsuperscript{2}\hspace{.2cm}
Pinaki Nath Chowdhury\textsuperscript{2}\\
Aneeshan Sain\textsuperscript{2}\hspace{.2cm}
Yi-Zhe Song\textsuperscript{2}\\
\textsuperscript{1} Department of Computer Science, University of Central Florida, United States.\\
\textsuperscript{2} SketchX, CVSSP, University of Surrey, United Kingdom.\\
{\tt\small Subhajit@ucf.edu; \{a.bhunia, s.koley, p.chowdhury, a.sain, y.song\}@surrey.ac.uk}\\
{\small \url{https://subhajitmaity.me/DYKp}}
\vspace{-7mm}
}

\begin{document}

\makeatletter
\let\@oldmaketitle\@maketitle
\renewcommand{\@maketitle}{\@oldmaketitle
  \centering
  \includegraphics[width=\linewidth]{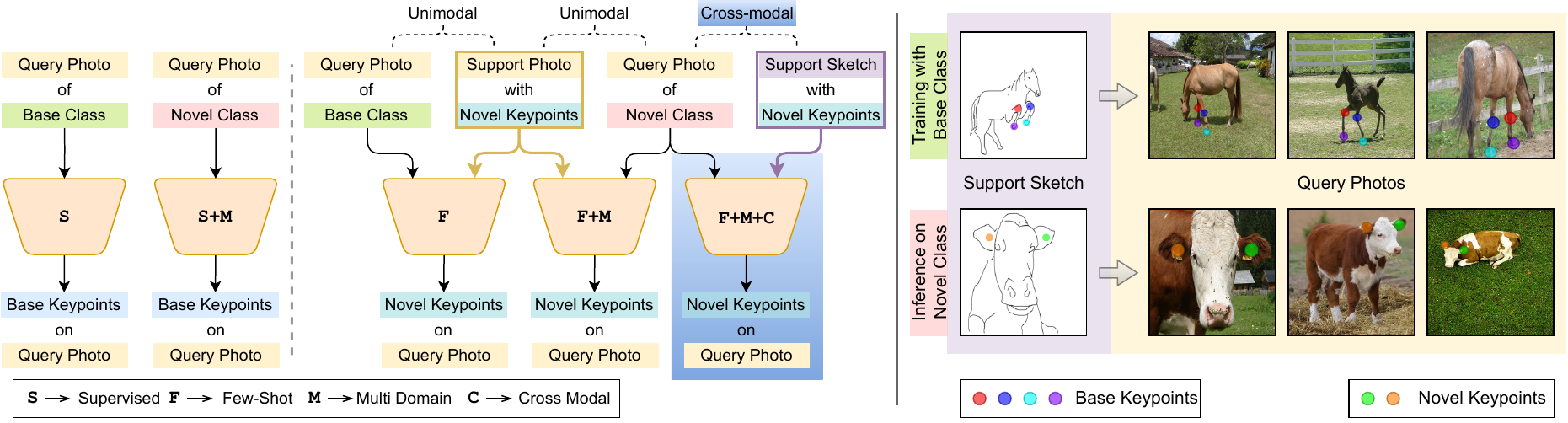}
  \vspace{-1em}
  \captionof{figure}{
  Keypoint detection~\cite{cao2021openpose}, usually approached in a supervised setting~\cite{fang2017rmpe}, along with cross-domain adaptation~\cite{cao2019cross}, can be set up in a few-shot paradigm~\cite{wei2021few} by typically learning to localize novel keypoints from limited annotated photos~\cite{he2023few}, also adapting to novel classes~\cite{Lu_2022_CVPR}. Unlike the existing works, we approach the problem in a cross-modal setup. \textit{(right)} The proposed few-shot framework adapts to localize novel keypoints on photos of unseen classes given a few annotated sketches.
  \label{fig:banner}
  \vspace{-0.03em}
  }
  \bigskip}
\makeatother

\maketitle

\begin{abstract}
Keypoint detection, integral to modern machine perception, faces challenges in few-shot learning, particularly when source data from the same distribution as the query is unavailable. This gap is addressed by leveraging sketches, a popular form of human expression, providing a source-free alternative. However, challenges arise in mastering cross-modal embeddings and handling user-specific sketch styles. Our proposed framework overcomes these hurdles with a prototypical setup, combined with a grid-based locator and prototypical domain adaptation. We also demonstrate success in few-shot convergence across novel keypoints and classes through extensive experiments.
\vspace{-1.5em}
\end{abstract}    
\section{Introduction}
\label{sec:intro}
\vspace{-0.25em}
Keypoint detection, a fundamental component in modern machine perception and visual understanding, has been integral to computer vision tasks since its inception~\cite{sun2013deep,cao2019cross,cao2021openpose,Yang_2023_CVPR}. Initially harnessed as \textit{distinctive} features~\cite{harris1988combined,lowe2004sift,bay2006surf}, keypoints have evolved from handcrafted designs to play crucial roles in contemporary pose estimation~\cite{cao2021openpose,Li_2021_ICCV} and landmark detection~\cite{thewlis2019unsupervised,wei2021few}. Their significance extends to applications such as localizing landmarks in faces~\cite{sun2013deep,thewlis2019unsupervised} and joints~\cite{hampali2022keypoint}, contributing to pose~\cite{cao2021openpose,fang2017rmpe,cheng2020higherhrnet,Yang_2021_ICCV} and action~\cite{Liu_2020_CVPR} understanding.

While existing approaches predominantly rely on heatmap regression or direct regression methods dependent on large annotated datasets~\cite{Yang_2021_ICCV,fang2017rmpe,newell2016stacked,cao2021openpose,carreira2016human,toshev2014deeppose}, few-shot keypoint learning remains a challenge. Current strategies, often repurposed from unsupervised or semi-supervised techniques, are constrained to specific image domains, limiting their \textit{applicability} to novel keypoints on unseen objects~\cite{Lu_2022_CVPR}. The scarcity of research addressing the challenge of unavailable source modality in few-shot keypoint learning, particularly for novel keypoints, highlights a significant gap in the current landscape.

Why choose \textit{sketches}? Sketches~\cite{hertzmann2020line} serve as a popular form of human expression and offer a viable alternative when the source data is limited in a few-shot setting. Unlike real images, human sketches, easily annotated, provide a practical solution for localizing novel keypoints in unseen classes. While generalized few-shot keypoint detection~\cite{Lu_2022_CVPR} must handle domain shifts and necessitates a few annotated examples, sketches present the opportunity to create a \textit{source-free} pipeline for practical implementations.

In the pursuit of utilizing sketches as an alternative support modality for a source-free few-shot keypoint detection approach, several distinct challenges emerge. First, mastering \textit{cross-modal} keypoint embeddings proves formidable due to inherent disparities between these embeddings and joint sketch-photo embedding as per conventional cross-modal learning. Second, encoding images within this joint feature space may lack direct \textit{correspondence} for specific keypoint embeddings, adding intricacy to the task. Third, the untested nature of few-shot capability within this cross-modal feature space introduces additional complexity. Simultaneously, sketches, as expressive forms of human representation rather than precise depictions of photorealism, introduce variability through user-specific styles, necessitating effective generalization across diverse styles to facilitate efficient keypoint learning.

To tackle these challenges, we adopt a prototypical setup~\cite{snell2017prototypical}, constructing keypoint \textit{prototypes} by pooling embeddings from encoded support feature maps. These prototypes are then correlated with query feature maps, generating rich keypoint descriptors. The subsequent use of the Grid Based Locator (GBL)~\cite{Lu_2022_CVPR} allows for the localization of keypoints from the rich descriptor through direct regression. To bridge the sketch-photo domain gap at the keypoint level, we introduce prototypical domain adaptation~\cite{tanwisuth2021a}, which addresses the disparity between support prototypes and query keypoint embeddings pooled from query feature maps. Managing user style diversity is deemed crucial, and therefore, we maximize the similarity between corresponding keypoint embeddings extracted from different sketch versions. Each version represents a single object-level photo and exhibits varying styles.

Our contributions include \textit{(a)} the introduction of a \textit{sketch-based prototypical keypoint detection} framework, \textit{(b)} the pioneering exploration of \textit{source-free} few-shot keypoint detection, and \textit{(c)} extensive experimentation and analysis establishing the proposed method as a successful solution for few-shot convergence on novel keypoints in novel classes within a \textit{source-free cross-modal} setup.
\vspace{-0.5em}

\vspace{-0.25em}
\section{Related Works}
\label{sec:related}
\vspace{-0.25em}
\keypoint{Sketch for Visual Understanding:}
Having a surprisingly similar \textit{interpretation} as real objects for human perception~\cite{hertzmann2020line}, sketch~\cite{chowdhury2023democratising} and line drawings~\cite{li2019photo,xsoria2020dexined,li2019im2pencil} have been a \textit{popular} mode of expression, driving applications of \textit{hand-drawn sketches} in multiple tasks including retrieval~\cite{bhunia2021more,bhunia2020sketch,sain2023exploiting,sain2023clip,chowdhury2022partially,yu2016sketch}, generation~\cite{ghosh2019interactive,koley2023picture}, in-painting~\cite{xie2021exploiting}, segmentation~\cite{hu2020sketch}, 3D modeling~\cite{zhang2021sketch2model}, and augmented reality~\cite{yan2020interactive}. In particular, sketch-based image retrieval~\cite{collomosse2019livesketch,dutta2019semantically,sain2022sketch3t,sain2021stylemeup,shen2018zero,yelamarthi2018zero} involving various deep networks~\cite{bhunia2020sketch,collomosse2019livesketch,ribeiro2020sketchformer} to learn a \textit{cross-modal embedding space} has been immensely popular among recent research directions due to its significant commercial importance. Due to the popularity of sketch as a \textit{barrier-free} mode of human interaction across ethnic and social diversity, it has been used to guide object detection~\cite{chowdhury2023can}, saliency exploration~\cite{bhunia2023sketch2saliency}, video synthesis~\cite{li2021deep}, editing~\cite{yang2020deep}, and other computer vision problems~\cite{chowdhury2023scenetrilogy,xu2022deep}. Attributing to the \textit{cognitive} and \textit{creative} potential~\cite{ge2020creative} of humane intelligence, sketch has been used for Pictionary-styled gaming~\cite{bhunia2020pixelor}. With all these diverse directions, the research community has explored the exploitation of both representative~\cite{wang2021sketchembednet} and discriminative~\cite{bhunia2022doodle,chowdhury2023can} properties of sketch images. A more detailed exploration of sketches for various visual understanding tasks and their comparative analysis with several future directions are given in~\cite{BHUNIAAYANKUMAR2022TPoS}. In this work, we particularly explore the \textit{representative} property of the sketch in a \textit{localized context} at a cross-modal latent for learning distinct keypoints in real photos.

\keypoint{Few-Shot Learning:}
Few-shot learning~\cite{wang2020generalizing} relates to meta-learning~\cite{yin2018adversarial} closely, delving into strategies promoting \textit{faster convergence} with \textit{fewer training samples} and is approached by a learning-to-learn~\cite{lake2015human} approach. From an early age meta-learning~\cite{weng2018metalearning} usually involved models~\cite{ramalho2018adaptive}, either specifically designed to use external memory buffers~\cite{graves2014neural,weston2014memory,santoro2016meta} and fast weights~\cite{munkhdalai2017meta}, or strategies~\cite{zhou2019learning,Rusu2018MetaLearningWL} designed to imitate and manipulate optimization~\cite{ravi2016optimization,Yao2020AutomatedRM,Mishra2018ASN,Qiao2018FewShotIR} and gradients~\cite{Finn2017ModelAgnosticMF,Nichol2018OnFM}. Irrespective of strategies like context adaptation~\cite{Zintgraf2018CAMLFC}, cross-modal setup~\cite{Xing2019AdaptiveCF,Schwartz2019BabyST,Peng2019FewShotIR}, Bayesian~\cite{Kim2018BayesianMM}, Low-shot~\cite{qi2018low,hariharan2017low,wang2018low,gao2018low}, or the recent metric learning approaches~\cite{cai2018memory,allen2019infinite,lifchitz2019dense,Yoon2019TapNetNN,oreshkin2018tadam}, modern few-shot learning uses a \textit{support} set of a few \textit{unseen} examples to maximize task-level performance on the \textit{query} set. Among metric-based approaches~\cite{koch2015siamese,vinyals2016matching,sung2018learning}, the prototypical network~\cite{snell2017prototypical} and improvements~\cite{Li_2019_ICCV,Li2019RevisitingLD} became popular recently. Few-shot strategies, primarily curated for classification to distinguish \textit{class-specific semantics}, cannot directly cater to keypoint detection due to significantly different task spaces. In this context, prototypes~\cite{snell2017prototypical}, carrying rich \textit{representative} features from \textit{semantically distinct keypoints}, are suitable for the few-shot keypoint learning regimes~\cite{Lu_2022_CVPR}.

\keypoint{Keypoint Detection:}
Ever since its inception~\cite{harris1988combined,lowe2004sift,bay2006surf}, keypoint detection has been one of the major research trends in computer vision. The advent of deep learning~\cite{sun2013deep,cao2017realtime} has propelled the strategies and the usefulness of keypoint detection. Although being motivated by object detection~\cite{ren2015faster,cai2018cascade} and semantic segmentation~\cite{he2017mask} research, keypoint detection has advanced object detection by rethinking it from keypoints~\cite{law2018cornernet,zhou2019bottom,zhou2019objects}. However, it has been traditionally used for identifying landmarks in the face~\cite{sun2013deep,thewlis2019unsupervised} and joints~\cite{hampali2022keypoint}, specifically to understand the skeletal structure or pose~\cite{cao2021openpose, Li_2021_ICCV} of both living~\cite{cao2019cross} and non-living objects~\cite{Yang_2023_CVPR}. Keypoint detection is typically solved either by employing a \textit{heat-based localization} strategy~\cite{Yang_2021_ICCV,fang2017rmpe,newell2016stacked,browatzki20203fabrec,sun2013deep}, or using \textit{direct regression}~\cite{carreira2016human,toshev2014deeppose}. The heatmap-based strategies, specifically the ones for pose estimation, can be further classified into \textit{top-down} approaches~\cite{fang2017rmpe,sun2019deep} and \textit{bottom-up} approaches~\cite{cao2021openpose,cheng2020higherhrnet,newell2016stacked,LuoSWAHR}. Keypoint detection research towards cross-domain adaptation~\cite{cao2019cross}, shape deformation aware pose translation~\cite{li2020deformation}, self-supervision~\cite{novotny2018self,jakab2020self}, contrastive learning~\cite{Bai_2023_WACV}, semi-supervision~\cite{dong2019teacher,moskvyak2020semi,qian2019aggregation,mathis2018deeplabcut,honari2018improving} and unsupervised pipelines~\cite{jakab2018unsupervised,lorenz2019unsupervised,thewlis2019unsupervised} were also seen. The initial few-shot approaches~\cite{browatzki20203fabrec,yao2021one} explored semi-supervised or few-shot adaptation~\cite{wei2021few} following self-supervised pre-training, without leaving any room for cross-modal adaptation. While some strategies~\cite{lu2024detect,sun2024uniap,ge2021metacloth,lu2023saliency} adapt to novel keypoints, exploiting shape-level~\cite{he2023few,wimmer2024back} or location-level~\cite{Lu_2022_CVPR} uncertainty, and additional textual guidance~\cite{lu2024openkd}, they inherently cannot address the domain gap between sketches to learn from and photos to localize on.

\begin{figure*}[t]
    \centering
    \includegraphics[width=\linewidth]{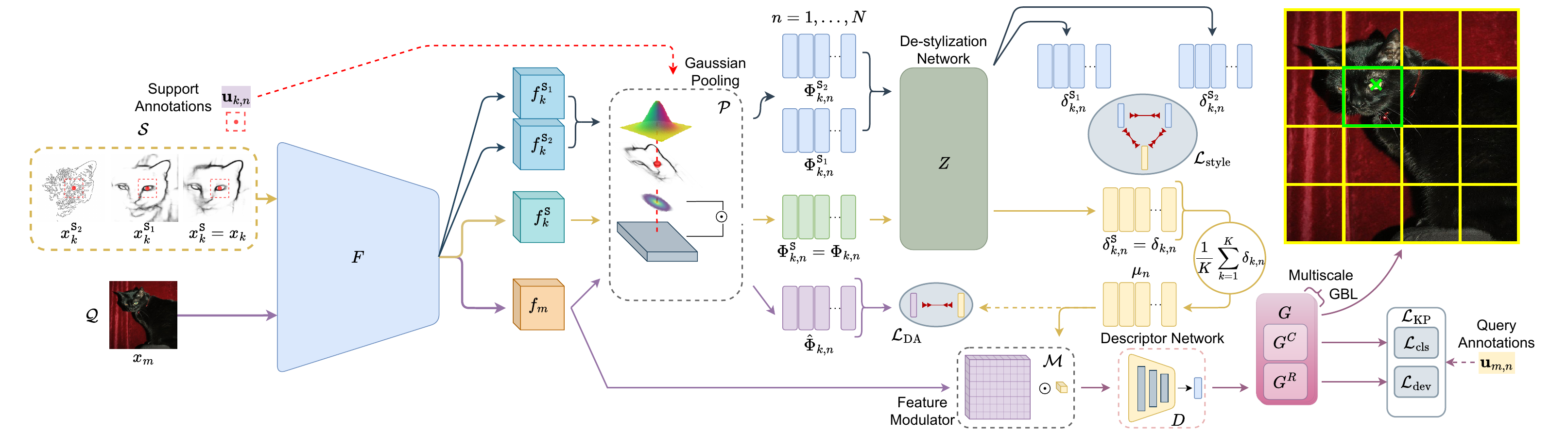}
    \caption{Overview of the proposed few-shot key-point detection framework that processes sketches or edgemaps in the support set and photos in the query set. It employs an encoder to extract deep feature maps followed by the derivation of keypoint embeddings through Gaussian Pooling. Support keypoint prototypes are constructed by averaging keypoint embeddings after disentangling style information through the de-stylization network. Support-query correlation is performed by a point-to-point multiplication of prototype and a query feature map, and subsequently a descriptor network formulates a query descriptor, which is used for localization by the GBL module.
    \vspace{-2em}
    }
    \label{fig:Architecture}
    \vspace{-1em}
\end{figure*}
\keypoint{Sketch as an Alternate to Photos:}
Multi-modal learning~\cite{wang2020makes} has been a recent trend to boost performance~\cite{lin2023multimodality} and perform semi-supervision~\cite{bhunia2021more}, and cross-modal~\cite{zhen2019deep,li2019cross} adaptations. In this regard, sketch~\cite{yu2016sketch,sangkloy2016sketchy} became a popular choice for cross-modal~\cite{bhunia2020sketch,sain2021stylemeup} counterparts of object detection~\cite{chowdhury2023can}, saliency recognition~\cite{bhunia2023sketch2saliency}~\etc Despite multimodal~\cite{pahde2021multimodal,huang2019acmm,Xing2019AdaptiveCF} approaches for \textit{few-shot} learning included sketch~\cite{bhunia2022doodle}, keypoints regime remained unexplored. Sketch~\cite{hertzmann2020line} being easy to achieve using a \textit{few strokes} remains a potential alternative for photos in particular scenarios,~\eg~when the photo collection is hazardous or the subject is a rare species of bird or animal, or when the photo cannot be used due privacy and ethical constraints or, particularly when variability in object pose and perspective becomes a big challenge by occluding keypoints.
\vspace{-0.75em}
\section{Methodology}
\label{sec:method}
\vspace{-0.25em}
\keypoint{Overview:}
In the context of few-shot keypoint detection, we have access to relevant dataset $\mathcal{D} = \{x^{\texttt{P}}_i, y_i\}$ with keypoint annotations of multiple species or objects where $x^{\texttt{P}}_i \in \mathcal{I}^{3 \times H \times W}$ is a standard RGB image and the annotation $y_i = \{\mathbf{u}_n, \mathbf{v}_n\}^N_{n=1}$ where $\mathbf{u} \in \mathbb{R}^2$ denotes coordinate location $\texttt{x},\texttt{y} \in [-1,1]$ of a keypoint, and visibility $\mathbf{v} \in \{0,1\}$ specifies if the keypoint is visible in the photo $x^{\texttt{P}}_i$, for each of the $N$ pre-defined keypoints. Due to the lack of a sketch-photo dataset with keypoint-level annotation, we use off-the-shelf edge detectors~\cite{su2021pixel,xie2015holistically,canny1986computational} to generate multiple versions of edgemaps from each photo $x^{\texttt{P}}_i$, considering them as equivalent to sketch data. In particular, we consider PiDiNet~\cite{su2021pixel} as the primary edge detector to generate $x^{\texttt{S}}_i$, and HED~\cite{xie2015holistically} and Canny~\cite{canny1986computational} as additional detectors providing \textit{style diversity}, respectively producing $x^{\texttt{S}_1}_i$ and $x^{\texttt{S}_2}_i$ against any real photo $x^{\texttt{P}}_i$. All the edgemaps are standard RGB images $\mathcal{I}^{3 \times H \times W}$. Thus, it is formulated as an N-way K-shot learning problem to detect $N$ keypoints on query photos using $K$ annotated sketches. The \textit{support} set $\mathcal{S} = {\{x_k, y_k\}}^K_{k=1}$ and \textit{query} set $\mathcal{Q} = {\{x_m, y_m\}}^M_{m=1}$ contains $K$ and $M$ distinct samples ${\{x_i, y_i\}}$ respectively from $\mathcal{D}$ where $x_k = x^{\texttt{S}}_k$ and $x_m = x^{\texttt{P}}_m$.
\subsection{Few-Shot Framework}
\label{sec:method_framework}
The proposed few-shot framework closely follows Lu~\etal~\cite{Lu_2022_CVPR} by building prototypes~\cite{snell2017prototypical} for all $N$ keypoints from support set $\mathcal{S}$, satisfying the requirements of keypoint localization, which is achieved by direct regression using a Grid-based Locator (GBL)~\cite{Lu_2022_CVPR} catering it towards sketch data (\cref{sec:method_framework_GBL}). The proposed architecture (\cref{fig:Architecture}) uses an image encoder $F$ to extract feature maps from the support edgemaps and the query photo. Subsequently, given the support keypoint locations, a keypoint extractor $\mathcal{P}$ extracts keypoint embeddings from support features, followed by computing support prototypes, which correlate to the query features using a feature modulator $\mathcal{M}$. Feature descriptors are obtained from correlated feature maps through a descriptor network~\cite{Lu_2022_CVPR} $D$ and lastly, the GBL modules (grid classifier $G^C$ and regressor $G^R$) operating in multi-scale localizes the keypoints. The system is trained with keypoints loss $\mathcal{L}_{\text{KP}}$. Additionally, a \textit{domain adaptation} strategy~\cite{tanwisuth2021a} optimizes a domain loss $\mathcal{L}_{\text{DA}}$ discussed in \cref{sec:method_da} and a novel \textit{de-stylization} network $Z$ utilizing an attention mechanism~\cite{dai2021attentional} maximizes information across diverse edgemap styles using $\mathcal{L}_{\text{style}}$ as elaborated in \cref{sec:method_style}. Moreover, auxiliary keypoints and tasks improve performance using auxiliary losses $\mathcal{L}_{\text{KP-aux}}$, $\mathcal{L}_{\text{DA-aux}}$, and $\mathcal{L}_{\text{style-aux}}$ as given in \cref{sec:method_aux}. The total loss is in \cref{equ:loss_total} where $\lambda_{\text{KP}}$, $\lambda_{\text{DA}}$ and $\lambda_{\text{style}}$ are corresponding loss scaling factors.
\vspace{-0.5em}
{\small
\begin{equation}
\label{equ:loss_total}
\begin{aligned}
    \mathcal{L}_{\text{total}} = \lambda_{\text{KP}}(\mathcal{L}_{\text{KP}} + \mathcal{L}_{\text{KP-aux}}) + \lambda_{\text{DA}}(\mathcal{L}_{\text{DA}} + \mathcal{L}_{\text{DA-aux}})\\
    + \lambda_{\text{style}}(\mathcal{L}_{\text{style}} + \mathcal{L}_{\text{style-aux}})
\end{aligned}
\vspace{-0.25em}
\end{equation}}

\subsubsection{Support Prototypes \& Query Correlation}
\label{sec:method_framework_proto}
\keypoint{Image Encoder:}
The image encoder $F: \mathcal{I}^{3 \times H \times W} \rightarrow \mathbb{R}^{c \times h \times w}$ is an essential component of the framework that takes an input $x_i$, \ie either a support sketch or edgemap $x^{\texttt{S}}_k$ or a query photo $x^{\texttt{P}}_m$, and encodes it to convolutional feature map $f_i$. Meaning, for each of the $K$ edgemaps $x_k \in \mathcal{S}$ and for each of the $M$ photos $x_m \in \mathcal{Q}$, the encoder $F$ generates support feature $f_k$ and query feature $f_m$ respectively.

\keypoint{Support Prototypes:}
The keypoint detection task inherently demands the learning of keypoint embeddings $\Phi_{k,n} \in \mathbb{R}^c$. So, we use $\mathcal{P}$ operator for extracting embedding $\Phi_{k,n} = \mathcal{P}(f_k, \mathbf{u}_{k,n})$ for keypoint $n$ in $x_k \in \mathcal{S}$. $\mathcal{P}$ is realized by Gaussian pooling~\cite{Lu_2022_CVPR} in \cref{equ:gauss_pool} to encode \textit{sufficiently discriminative} local context \textit{without} any hard local boundary at the corresponding ground-truth location $\mathbf{u}_{k,n} \in y_k$, where $f_k[\mathbf{x}]$ refers to the $\mathbb{R}^c$ vector from $f_k$ indexed at location $\mathbf{x}$.
\vspace{-0.5em}
{\small
\begin{equation}
\label{equ:gauss_pool}
\vspace{-0.25em}
\begin{aligned}
    \mathcal{P}(f_k, \mathbf{u}_{k,n}) = \sum_{\mathbf{x}} \exp\bigg(\frac{-\|\mathbf{x} - \mathbf{u}_{k,n}\|_2^2}{2\xi^2}\bigg) \cdot f_k[\mathbf{x}]
\end{aligned}
\vspace{-0.5em}
\end{equation}}
Each $\Phi_{k,n}$ from support $x_k$ is mapped to a \textit{style-agnostic} embedding $\delta_{k,n} \in \mathbb{R}^c$ enriched with \textit{mutual information across styles} (\cref{sec:method_style}) using de-stylization network $Z$.

Following the standard prototypical network~\cite{snell2017prototypical}, we compute each support keypoint prototype $\mu_n:n=1, \ldots, N$ as a mean of the visible de-stylized support keypoint embeddings $\delta_{k,n}$ across the support set $\mathcal{S}$ as per \cref{equ:prototype}. The visibility of the prototype $\mu_n$ is \texttt{true} if $\mathbf{v}_{k,n} = 1$ for at least one $x_k \in \mathcal{S}$. The loss uses it to not penalize the model for predicting keypoints absent in $\mathcal{S}$.
{\small
\vspace{-0.5em}
\begin{equation}
\label{equ:prototype}
\vspace{-0.5em}
\begin{aligned}
    \mu_n = \frac{1}{K} \sum_{k=1}^K \delta_{k,n}
\end{aligned}
\vspace{-0.25em}
\end{equation}}

\keypoint{Support-Query Correlation:}
While the original prototypical network~\cite{snell2017prototypical} used a distance metric to assign class probabilities for the classification task, the keypoint localization is inherently different from it and requires local-level feature correspondence for each keypoint. Hence, support prototypes $\mu_n:n=1, \ldots, N$ for all the keypoints are separately correlated with query feature $f_m$ encoded from query photo $x_m \in \mathcal{Q}$ with a feature modulator~\cite{Lu_2022_CVPR} $\mathcal{M}$.
\vspace{-0.25em}
{\small
\begin{equation}
\label{equ:correlate}
\begin{aligned}
    \mathcal{A}_{m,n} = \mathcal{M}(f_m, \mu_n) \qquad\qquad \mathcal{A}_{m,n}[\mathbf{x}] = f_m[\mathbf{x}] \odot \mu_n
\end{aligned}
\vspace{-0.25em}
\end{equation}}

$\mathcal{M}$ is defined by elementwise multiplication of prototype $\mu_n$ and query feature $f_m$ at every location index $\mathbf{x}$ of $f_m$ for each of the $N$ keypoints, as in \cref{equ:correlate} and generates correlated feature $\mathcal{A}_{m,n} \in \mathbb{R}^{c \times h \times w}$ corresponding to $x_m$.

Following, Lu~\etal~\cite{Lu_2022_CVPR} a \textit{descriptor} network $D:\mathcal{A}_{m,n} \rightarrow \Psi_{m,n}$ refines and encodes the positional information in the query photo $x_m$ from the correlated feature maps $\mathcal{A}_{m,n}$ to descriptors $\Psi_{m,n} \in \mathbb{R}^d$ for all $N$ keypoints.
\subsubsection{Keypoints Localization}
\label{sec:method_framework_GBL}
\keypoint{Overview:}
Grids~\cite{roth2009classifier,najibi2016g,redmon2016you}, being used for more than a decade for detection~\cite{redmon2016you,Redmon_2017_CVPR,redmon2018yolov3} tasks, were extended to keypoint localization~\cite{Lu_2022_CVPR} using a grid regression by modeling keypoint uncertainty. Inspired by the same, we use a direct regression in our GBL without any notion of uncertainty, as it becomes daunting due to the \textit{sparse} nature of sketches.

We dismantle the localization task of GBL into two sub-problems, \textit{grid classification} and \textit{local grid offset regression}, to restrict the error margin to a small patch. To be precise, to localize keypoint $n$ in a query photo $x_m$ in multi-scale, a set of $S$ grid-scales $L = \{L_1, \ldots, L_S\}$ is considered and each grid-scale $L_i \in L$ is used to divide $x_m$ into $L_i \times L_i$ patches across height and width, of size $3 \times \frac{H}{L_i} \times \frac{W}{L_i}$ each. The patches are labelled as the set $l_i = \{0,1,2,\ldots,({L_i}^2 - 1)\}$ in \textit{left-to-right} and \textit{top-to-bottom} order. The ground-truth $\mathbf{u}_n$ in the corresponding $y_m$, in the range $[-1,1]$, is converted to the grid map label $\mathbf{u}^C_n$ and offset $\mathbf{u}^R_n$ as per \cref{equ:gbl_gt}. Formally, $\mathbf{t}$ denotes the ground-truth coordinates $\texttt{x},\texttt{y} \in \mathbf{u}_n$ normalized in $[0,L_i]$, and $\mathbf{z}$ refers to the $\texttt{x},\texttt{y}$ indexing of the grids $l_i$. Next, the grid index $\mathbf{z} = \{\mathbf{z}_{\texttt{x}}, \mathbf{z}_{\texttt{y}}\}$ is converted to actual grid label $\mathbf{u}^C_n \in l_i$ and the grid offset $\mathbf{u}^R_n$ is the deviation from the center,~\ie location inside the grid $\mathbf{u}^C_n$ normalized in $[-1,1]$.

\vspace{-1.25em}
{\small
\begin{equation}
\label{equ:gbl_gt}
\begin{aligned}
    \mathbf{t} = (\mathbf{u}_n/2 + 0.5) * L_i \qquad & \mathbf{z} = \lfloor \max (0, \min (\mathbf{t}, 1)) \rfloor\\
    \mathbf{u}^C_n = \mathbf{z}_{\texttt{y}} \cdot L_i + \mathbf{z}_{\texttt{x}} \qquad & \mathbf{u}^R_n = 2(\mathbf{t} - \mathbf{z} -0.5)
\end{aligned}
\vspace{-0.5em}
\end{equation}}

The submodules of our version of GBL, the grid classifier $G^C$ and the grid regressor $G^R$ try to predict $\mathbf{u}^C_n$ and $\mathbf{u}^R_n$ respectively for localizing keypoints. The total loss is given $\mathcal{L}_{\text{KP}} = \mathcal{L}_{\text{cls}} + \mathcal{L}_{\text{dev}}$ which optimizes a completely different objective compared to Lu~\etal~\cite{Lu_2022_CVPR}. All the loss calculations and predictions for keypoint $n$ are considered only if it is visible in $x_m$ and has a visible support prototype $\mu_n$.

\keypoint{Grid Classifier:}
The grid classifier $G^C$ learns to estimate the probability distribution $\hat{\mathbf{u}^C_n} = p(l|\Psi_{m,n})$ of keypoint $n$ being located in patch $l \in l_i$ from the query \textit{keypoint descriptor} $\Psi_{m,n}$ using \texttt{softmax} on a linear layer as described in \cref{equ:gbl_clsprob}, with an output space $\mathbb{R}^{{L_i}^2}$ for each keypoint $n$. This specifically resolves the dependency of grid regressor by providing it with a smaller scope to localize.
\vspace{-0.5em}
{\small
\begin{equation}
\label{equ:gbl_clsprob}
\begin{aligned}
    p(l|\Psi_{m,n}) = \frac{\exp(G^C(\Psi_{m,n})_l)}{\sum_{l' \in l_i}\exp(G^C(\Psi_{m,n})_{l'})}
\end{aligned}
\vspace{-0.25em}
\end{equation}}
While Lu~\etal~\cite{Lu_2022_CVPR} considers grid-level uncertainty using log-likelihood, we opted to train $G^C$ with the much simpler cross-entropy loss $\mathcal{L}_{\text{cls}}$, as given in \cref{equ:gbl_loss_cls}, leveraging the one-hot encoded ground-truth $\mathds{1}(\mathbf{u}^C_n)$.

\vspace{-0.5em}
{\small
\noindent
\begin{minipage}[b]{0.57\linewidth}
\begin{equation}
\label{equ:gbl_loss_cls}
    \mathcal{L}_{\text{cls}} = - \sum \log(\hat{\mathbf{u}^C_n}) \odot \mathds{1}(\mathbf{u}^C_n)
\vspace{-1em}
\end{equation}
\end{minipage}
\noindent
\begin{minipage}[b]{0.41\linewidth}
\begin{equation}
\label{equ:gbl_loss_dev}
    \mathcal{L}_{\text{dev}} = {\|\hat{\mathbf{u}^R_n} - \mathbf{u}^R_n\|}_1
\vspace{-0.75em}
\end{equation}
\end{minipage}
\vspace{0.75em}
}

\keypoint{Grid Regressor:}
The grid regressor $G^R$ tries to predict the offset values $\hat{\mathbf{u}^R_n} = G^R(\Psi_{m,n})$ against the relevant grid $l$ (from ground-truth $\mathbf{u}^C_n$ during training, and from grid classifier prediction $\max(\hat{\mathbf{u}^C_n})$ during inference) for keypoint $n$ from the corresponding query \textit{keypoint descriptor} $\Psi_{m,n}$ using a linear layer with output space $\mathbb{R}^{2}$. Unlike modeling the offset as a sample of multivariate Gaussian distribution in Lu~\etal~\cite{Lu_2022_CVPR}, our $G^R$ uses an $l_1$ loss $\mathcal{L}_{\text{dev}}$ as in \cref{equ:gbl_loss_dev}.

\keypoint{Prediction:}
The final prediction for keypoint $n$ in query $x_m$ is computed from the predicted grid label probability $\hat{\mathbf{u}^C_n}$ and grid offset $\hat{\mathbf{u}^R_n}$. Considering the predicted grid label as $\max (\hat{\mathbf{u}^C_n})$, for any grid scale, the prediction is computed by adding $\hat{\mathbf{u}^R_n}$ to its center. The final prediction is given as a mean across all $S$ grids in $L$.
\subsection{Optimizing Sketch-Photo Domain Gap}
\label{sec:method_da}
From the problem definition, the sketch-photo domain gap remains the biggest challenge as no sketch $x_k \in \mathcal{S}$ has \textit{content-level pairing} with any of the photo $x_m \in \mathcal{Q}$. Considering this, we seek guidance from modern domain adaptation~\cite{tanwisuth2021a} that particularly implements a prototypical framework~\cite{snell2017prototypical} to optimize \textit{transport loss} pulling source prototypes and target samples closer. However, as keypoint detection differs significantly from the classification at the task level, the transport loss~\cite{tanwisuth2021a} cannot be used naively. Moreover, the unsupervised setting~\cite{tanwisuth2021a} practically ignores the keypoint-level class information available.

Bi-directional transport loss~\cite{tanwisuth2021a} uses source and target data along with a common prototype in an unsupervised setting. As the cross-modal task space needs domain adaptation at the \textit{keypoint-level} instead of \textit{feature-level}, we consider support keypoint embeddings $\phi_{k,n}$ and extract similar keypoint embeddings $\hat{\Phi}_{m,n}$ from query photo $x_m$.

Transport loss~\cite{tanwisuth2021a} uses a \textit{point-to-point} cost $\mathbf{C}(\mu_n, \hat{\Phi}_{m,n}) = {\|\mu_n - \hat{\Phi}_{m,n}\|}_2$ between prototypes and query keypoints which is realized by $l_2$ distance. It also relies on a cosine similarity score between $\mu_n$ and $\hat{\Phi}_{m,n}$ as an \textit{unnormalized class likelihood} measure to adjust the cost $\mathbf{C}$, which is adapted to a normalized distance-based similarity measure $\mathbf{Sim}(\mu_n, \hat{\Phi}_{m,n}) = \exp{(-{\|\mu_n - \hat{\Phi}_{m,n}\|}_2^2)}$ signifying the likelihood of a query keypoint embedding $\hat{\Phi}_{m,n}$ relating to corresponding support keypoint prototype $\mu_n$, to encourage transport guided by keypoint-level \textit{representative} information instead of \textit{discriminative} class information. See \S\hyperlink{./X_suppl.tex}{Suppl.}
{\small
\vspace{-1.25em}
\begin{equation}
\label{equ:proto_prob}
\begin{aligned}
    p(\hat{\mu}_n) = \exp{(-{\|\hat{\mu}_n - \mu_n\|}_2^2)}~\qquad\hat{\mu}_n = \frac{1}{M} \sum_{m=1}^M \hat{\Phi}_{m,n}
\end{aligned}
\vspace{-0.75em}
\end{equation}}
While iteratively updating the target distribution $p(\hat{\mu}_n)$ was necessary for an unsupervised setting, we have access to the information of query keypoint correspondence to the prototypes and thus, we convert this domain adaptation~\cite{tanwisuth2021a} paradigm to a supervised one using \cref{equ:proto_prob}. As the class-level probability distribution to relate the prototype $\mu_n$ to query embedding $\hat{\Phi}_{m,n}$ is not coherent with our task, we ignore the \texttt{softmax} operation from both the transport losses, resulting in reduction of two transport losses~\cite{tanwisuth2021a} to one:
\vspace{-0.5em}
{\small
\begin{equation}
\label{equ:loss_da}
\begin{aligned}
    \mathcal{L}_{\text{DA}} = \sum p(\hat{\mu}_n) \mathbf{C}(\mu_n, \hat{\Phi}_{m,n}) \mathbf{Sim}(\mu_n, \hat{\Phi}_{m,n})
\end{aligned}
\vspace{-0.5em}
\end{equation}}

\subsection{Learning Style-Agnostic Keypoints}
\label{sec:method_style}
\begin{figure}[t]
    \centering
    \includegraphics[width=\textwidth]{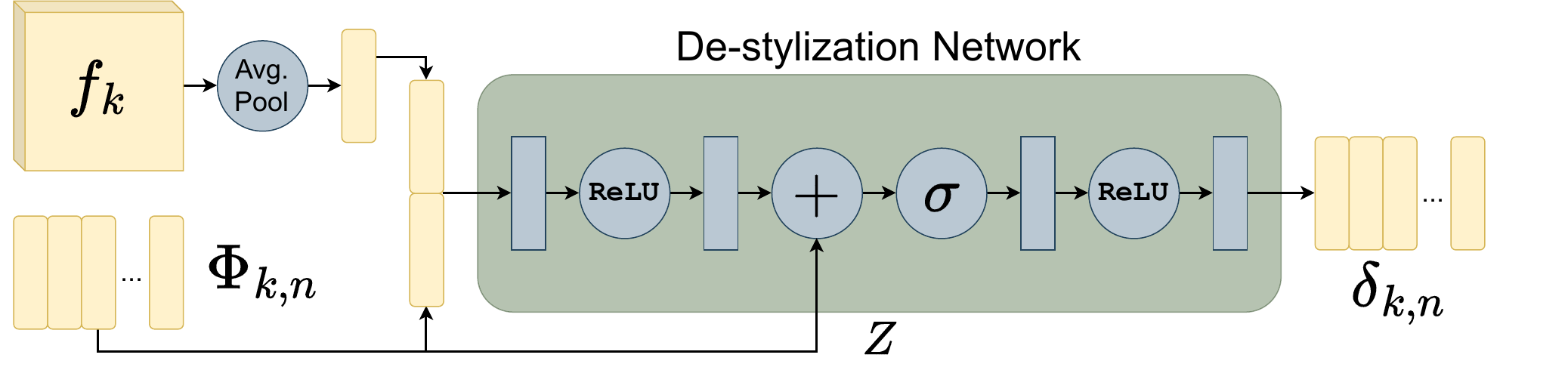}
    \vspace{-2em}
    \caption{
    The design of the de-stylization network $Z$ to disentangle the styles fusing global context to local keypoint embeddings.
    \vspace{-1.75em}
    }
    \label{fig:style}
\end{figure}

Sketch largely depends on the user's \textit{expressive intent and imagination} as opposed to \textit{photorealism} and has significant user-specific variance in terms of sparsity, information content, and style~\cite{sain2021stylemeup}. This style diversity impacts performance as the framework eventually learns dense information at the keypoint level, irrespective of sketch sparsity. So, we design a de-stylization network $Z$ that de-stylizes any support keypoint embedding $\Phi_{k,n}$ extracted from support edgemap $x_k$ to $\delta_{k,n} \in \mathbb{R}^c$. The off-the-shelf edge detectors~\cite{su2021pixel,xie2015holistically,canny1986computational} generate \textit{different instances} of edgemap $x^{\texttt{S}}_k$, $x^{\texttt{S}_1}_k$ and $x^{\texttt{S}_2}_k$ for the \textit{same} photo $x^{\texttt{P}}_k \in \mathcal{S}$ not only providing the style diversity but also aiding the encoder to learn dense features to tackle inherent sparsity of the sketch data. The design of the de-stylization network $Z$ illustrated in \cref{fig:style} is inspired by the \textit{multi-scale channel attention} module~\cite{dai2021attentional} that aggregates local and global context. (Design detailed in \S\hyperlink{./X_suppl.tex}{Suppl.}) The \textit{disentanglement} being targetted at the keypoint embeddings with the local context, $Z$ is devised to take the encoded feature $f_k^{\texttt{S}}$, $f_k^{\texttt{S}_1}$ and $f_k^{\texttt{S}_2}$ respectively, providing the notion of sparsity in a global scale along with keypoint embeddings $\Phi_{k,n}^{\texttt{S}}$, $\Phi_{k,n}^{\texttt{S}_1}$ and $\Phi_{k,n}^{\texttt{S}_2}$ for mapping to $\delta_{k,n}^{\texttt{S}}$, $\delta_{k,n}^{\texttt{S}_1}$ and $\delta_{k,n}^{\texttt{S}_2}$ respectively using an attention mechanism~\cite{dai2021attentional}.
Considering all the versions of edgemap $x^{\texttt{S}}_k$, $x^{\texttt{S}_1}_k$ and $x^{\texttt{S}_2}_k$ as supplements to various user styles with different level of abstraction and information content, $Z$ ideally should disentangle the same and map any keypoint \textit{similar} to each other. Taking $x^{\texttt{S}}_k$ as primary sketch equivalent for training, we have $f_k = f_k^{\texttt{S}}$, $\Phi_{k,n} = \Phi_{k,n}^{\texttt{S}}$ and $\delta_{k,n} = \delta_{k,n}^{\texttt{S}}$. Thus $\mathcal{L}_{\text{style}}$ is derived as: 
\vspace{-0.75em}
{\small
\begin{equation}
\label{equ:loss_style}
\begin{aligned}
    \mathcal{L}_{\text{style}} = \sum_n({\|\delta_{k,n} - \delta_{k,n}^{\texttt{S}_1}\|}_2 + {\|\delta_{k,n} - \delta_{k,n}^{\texttt{S}_2}\|}_2 +
    {\|\delta_{k,n}^{\texttt{S}_1} - \delta_{k,n}^{\texttt{S}_2}\|}_2)
\end{aligned}
\vspace{-0.75em}
\end{equation}}
\vspace{-2em}
\subsection{Auxiliary Keypoints Improve Performance}
\label{sec:method_aux}
Along with the main keypoints given by the annotations, we strategically \textit{interpolate} auxiliary keypoints~\cite{Lu_2022_CVPR} for the episodic training of our framework. Precisely, given two visible annotated main keypoints $n_1$ and $n_2$, and a relative position $t \in (0,1)$, interpolation $\mathcal{T}(t,(\mathbf{u}_{n_1}, \mathbf{u}_{n_2}))$ obtains an auxiliary keypoint at $t$ times the length on the line running from $n_1$ to $n_2$. Simply, auxiliary keypoints~\cite{Lu_2022_CVPR} are considered on the line interpolating a particular keypoint to another when both are visible. An off-the-shelf saliency detector~\cite{Wu_2019_ICCV} is used to see if an auxiliary keypoint is inside the saliency region of the object, determining its visibility.

While the generation strategy $\mathcal{T}$ is inspired by Lu~\etal~\cite{Lu_2022_CVPR}, our auxiliary keypoints serve a completely different set of objectives. Unlike using auxiliary keypoints, particularly to aid the uncertainty learning, we do not model any interaction between main and auxiliary keypoints. Instead, we use these additional keypoints to provide additional \textit{augmented} data for the components in our framework. To this end, we extend the existing objectives of localization, domain adaptation, and de-stylization to $T$ auxiliary keypoints by creating an auxiliary task of $T$ way keypoint learning and calculating $\mathcal{L}_{\text{KP-aux}}$, $\mathcal{L}_{\text{DA-aux}}$, and $\mathcal{L}_{\text{style-aux}}$ accordingly.
\vspace{-0.5em}
\section{Experiments}
\label{sec:expt}
\vspace{-0.5em}
\keypoint{Dataset:}
For the experiments, we have used the Animal Pose dataset~\cite{cao2019cross} consisting of 5 different species of animals (cat, cow, dog, horse, sheep) with a total of 4,666 images and 6,117 instances. The annotations contain 20 keypoints for each species, from which we used 17 (11 as \textit{base} for training and 6 as \textit{novel} for evaluation). All the ablation experiments are also done on this dataset.

Additionally, we consider Animal Kingdom dataset~\cite{Ng_2022_CVPR} with 33,099 photos of 850 different species of broadly 5 superclasses (mammal, amphibian, reptile, bird, fish), which we treat as image classes for our purpose. It contains 23 keypoint annotations for each photo, and keeping similarity, we use only 11 \textit{base} keypoints and 7 \textit{novel} keypoints.

\keypoint{Implementation Details:}
The entire framework is implemented in Python with PyTorch~\cite{paszke2019pytorch} and is trained on a 6GB Nvidia RTX 3060 mobile GPU. The encoder $F$ uses a Imagenet~\cite{deng2009imagenet} pre-trained ResNet50~\cite{he2016deep} backbone.
The descriptor network $D$ uses 3 \texttt{convolution} layers with \texttt{ReLU}~\cite{agarap1803deep} in sequence. The de-stylization network $Z$ follows \cref{fig:style}. The GBL components, $G^C$ and $G^R$ use a \texttt{linear} layer each, with an additional \texttt{softmax} for $G^C$.

We set $K=1$ shot and $M=5$ with $N=11$ base keypoints for training and $N=6$ novel keypoints for evaluation. The ablations also use the same setup unless otherwise specified. The input images $x_i$, for both sketch or edgemap $x_k$ and photo $x_m$ are resized to the height and width $H=W=384$ and thus the resulting feature map $f_i$ has height and width $h=w=12$. According to the encoder $F$ configuration, the feature map $f_i$ has $c=2048$ channels and the descriptor network $D$ allows it to produce query keypoint descriptors of dimension $d=4096$. The auxiliary keypoints use 6 pre-defined pairs from 11 base keypoints and set $t=\{0.25,0.5,0.75\}$ as interpolation points; meaning a maximum of 18 auxiliary keypoints can possibly exist for any $x_i$ during training. The multi-scale aspect of GBL is realized using 3 grid scales, $L = \{8, 12, 16\}$. The empirically determined hyper-parameters are $\xi=14$, $\lambda_{\text{KP}}=0.5$, $\lambda_{\text{DA}}=0.001$ and $\lambda_{\text{style}}=0.001$. The entire setup is trained for 80,000 iterations of episodes with Adam~\cite{kingma2014adam} optimizer with a learning rate of $0.0001$.

\begin{figure}[t]
\includegraphics[width=0.951\columnwidth]{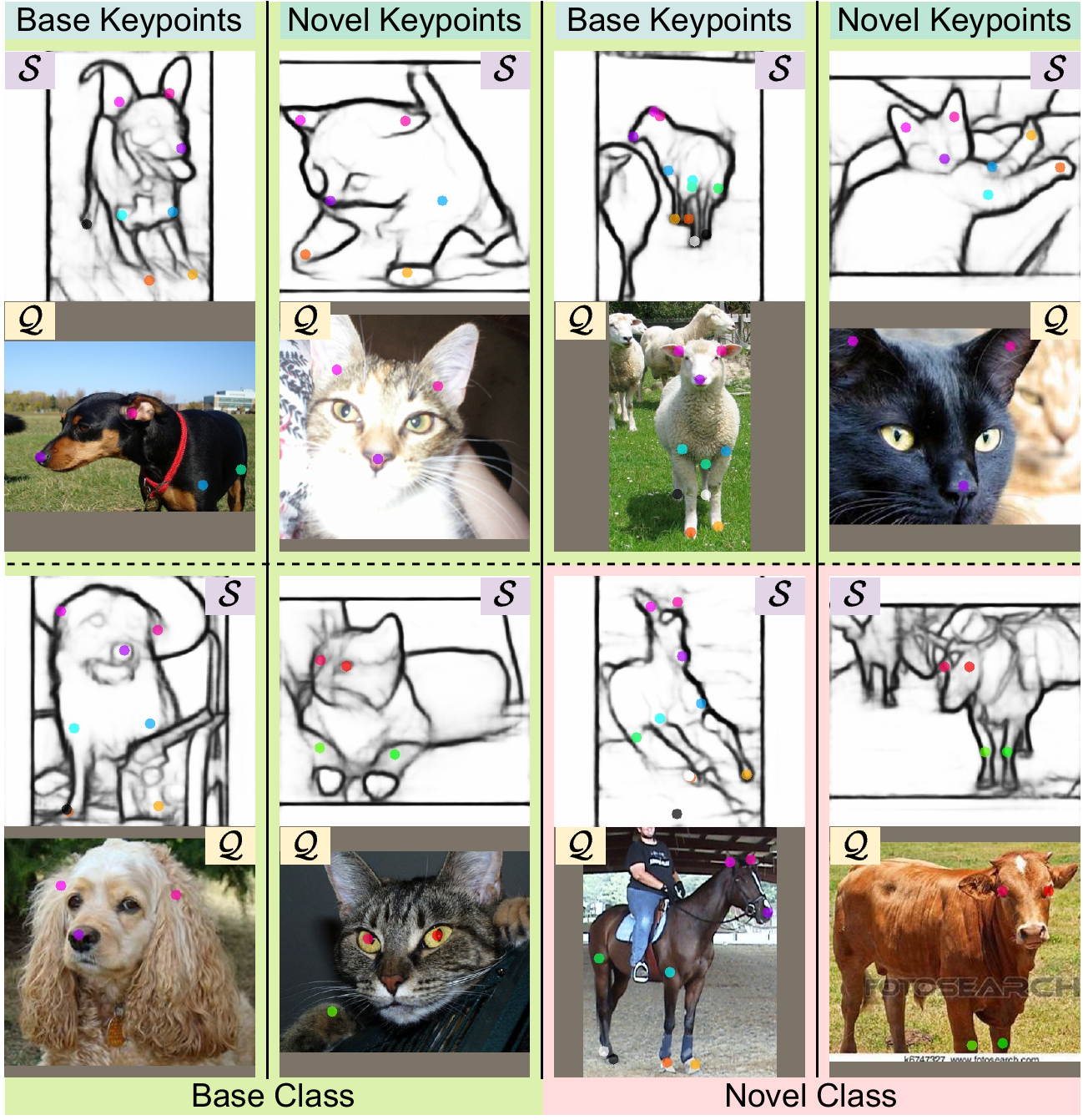}
\caption{Sample support ($\mathcal{S}$) and query ($\mathcal{Q}$) for training (\textit{top}) and evaluation (\textit{bottom}) for all evaluation settings.
\vspace{-3em}
}
\label{fig:sample_sets}
\vspace{-1.5em}
\end{figure}

\keypoint{Evaluation Protocol:}
We use percentage correct keypoints (PCK)~\cite{novotny2018self} with a margin $\tau=0.1$ as metric. The prediction is correct if within a range $\tau$ times the larger side of the object bounding box. We consider four different settings (see \cref{fig:sample_sets}) to analyze the performance of our framework. From the definition of few-shot keypoint detection, we evaluate performance with novel keypoints along with the base keypoints used in training for the base classes it is trained on, as well as never-before-seen classes. All the ablation studies are done for \textit{novel keypoints on unseen classes} of Animal Pose~\cite{cao2019cross} dataset unless otherwise specified. For the evaluation of a novel class, we trained the framework with episodes with data from the rest of the classes, and for seen classes, we used a random $7:3$ train-test split.

\keypoint{Competitors:}
Due to the lack of existing works on \textit{cross-modal} few-shot keypoint learning, we repurposed the next best alternative FSKD~\cite{Lu_2022_CVPR} to use edgemaps~\cite{su2021pixel} as support and photos as query, providing a comparable strategy, thanks to its robustness in detecting novel keypoints across significant differences in object classes.

We establish a vanilla baseline \textbf{B-Vanilla} using only the base framework (\cref{sec:method_framework}). For B-Vanilla, $Z$ is replaced by an identity function such that we have $\Phi_{k,n} = \delta_{k,n}$ and the additional edgemaps $x^{\texttt{S}_1}_k$ and $x^{\texttt{S}_2}_k$ are not used, setting $\mathcal{L}_{\text{style}} = 0$. We further establish a few incremental baselines. \textbf{B-DA} represents the usage of sketch-photo domain adaptation~\cite{tanwisuth2021a} on top of B-Vanilla. \textbf{B-Style} uses only style-agnostic keypoint learning presented in \cref{sec:method_style} on B-Vanilla. As \textbf{B-Full} uses both domain adaptation (\cref{sec:method_da}) and de-stylization (\cref{sec:method_style}), it provides the proposed model when used with auxiliary keypoints (\cref{sec:method_aux}).

\begin{table*}[t]
\caption{A quantitative comparison of the baseline and state-of-the-art strategies with the proposed framework in $K=1$ shot setting delineating the superiority of the proposed method in overall performance on Animal Pose~\cite{cao2019cross} and Animal Kingdom~\cite{Ng_2022_CVPR} datasets.
\vspace{-2em}
}
\centering
\scriptsize
\setlength{\tabcolsep}{6.4pt}
\begin{tabular}{lllcccccgcccccg}
\toprule
\multirow{2}{*}{Class}  & \multirow{2}{*}{Keypoints} & \multirow{2}{*}{Methods} & \multicolumn{6}{c}{PCK@0.1 on Animal Pose Dataset~\cite{cao2019cross}} & \multicolumn{6}{c}{PCK@0.1 on Animal Kingdom Dataset~\cite{Ng_2022_CVPR}} \\
\cmidrule{4-9} \cmidrule{10-15}
                        &                            &                          & Cat   & Cow   & Dog   & Horse & Sheep & \textbf{Mean}  & Mammal & Amphibian & Reptile & Bird & Fish & \textbf{Mean} \\
\midrule
\multirow{6}{*}{Seen}   & \multirow{3}{*}{Base}      & B-Vanilla                & 54.12 & 39.27 & 44.65 & 45.58 & 37.17 & 44.16          & 22.32   & 21.07      & 18.94    & 21.77 & 18.19  & 20.46 \\
                        &                            & FSKD~\cite{Lu_2022_CVPR} & 58.95 & 44.61 & 49.53 & 50.21 & 40.47 & 48.75          & 25.52   & 24.42      & 21.81    & 25.73 & 22.24  & 23.94 \\
                        &                            & \textbf{Proposed}        & \textbf{67.34} & \textbf{49.89} & \textbf{56.28} & \textbf{56.35} & \textbf{45.65} & \textbf{55.10} & \textbf{31.31}      & \textbf{29.93}      & \textbf{28.41}      & \textbf{30.88}      & \textbf{27.87}      & \textbf{29.68} \\
\cmidrule{2-15}
                        & \multirow{3}{*}{Novel}     & B-Vanilla                & 24.70 & 15.62 & 19.08 & 12.45 & 18.44 & 18.06          &  9.67   &  7.24      &  6.55    &  8.41 &  4.96  &  7.37 \\
                        &                            & FSKD~\cite{Lu_2022_CVPR} & 47.70 & 35.44 & 39.81 & 35.42 & 31.59 & 37.99          & 18.89   & 17.39      & 16.72    & 19.23 & 15.94  & 17.63 \\
                        &                            & \textbf{Proposed}        & \textbf{55.69} & \textbf{43.09} & \textbf{46.58} & \textbf{43.94} & \textbf{36.39} & \textbf{45.14} & \textbf{25.05}      & \textbf{23.27}      & \textbf{22.56}      & \textbf{24.04}      & \textbf{21.78}      & \textbf{23.34} \\
\midrule
\multirow{6}{*}{Unseen} & \multirow{3}{*}{Base}      & B-Vanilla                & 43.03 & 40.31 & 36.28 & 44.72 & 38.03 & 40.47          & 12.83   & 14.12      & 12.57    & 13.68 & 12.41  & 13.12 \\
                        &                            & FSKD~\cite{Lu_2022_CVPR} & 41.54 & 38.10 & 33.72 & 41.02 & 36.30 & 38.14          & 14.86   & 14.28      & 13.79    & 15.65 & 12.16  & 14.15 \\
                        &                            & \textbf{Proposed}        & \textbf{47.36} & \textbf{42.97} & \textbf{38.30} & \textbf{46.17} & \textbf{41.03} & \textbf{43.17} & \textbf{21.98}      & \textbf{20.15}      & \textbf{18.96}      & \textbf{21.52}      & \textbf{17.19}      & \textbf{19.96} \\
\cmidrule{2-15}
                        & \multirow{3}{*}{Novel}     & B-Vanilla                & 22.71 & 15.56 & 16.92 & 13.58 & 18.18 & 17.39          &  7.26   &  5.12      &  3.93    &  5.69 &  4.08  &  5.22 \\
                        &                            & FSKD~\cite{Lu_2022_CVPR} & 36.75 & 35.76 & 32.84 & 32.66 & 31.58 & 33.92          & 10.96   &  9.34      &  9.68    & 11.45 &  8.86  & 10.06 \\
                        &                            & \textbf{Proposed}        & \textbf{44.42} & \textbf{40.13} & \textbf{36.91} & \textbf{37.77} & \textbf{35.77} & \textbf{39.00} & \textbf{16.48}      & \textbf{14.62}      & \textbf{13.76}      & \textbf{15.91}      & \textbf{11.33}      & \textbf{14.42} \\
\bottomrule
\end{tabular}
\label{tab:results}
\vspace{-1em}
\end{table*}

\vspace{-0.25em}
\subsection{Performance Analysis}
\label{sec:expt_performance}
\vspace{-0.25em}
\begin{figure}[t]
\includegraphics[width=\textwidth]{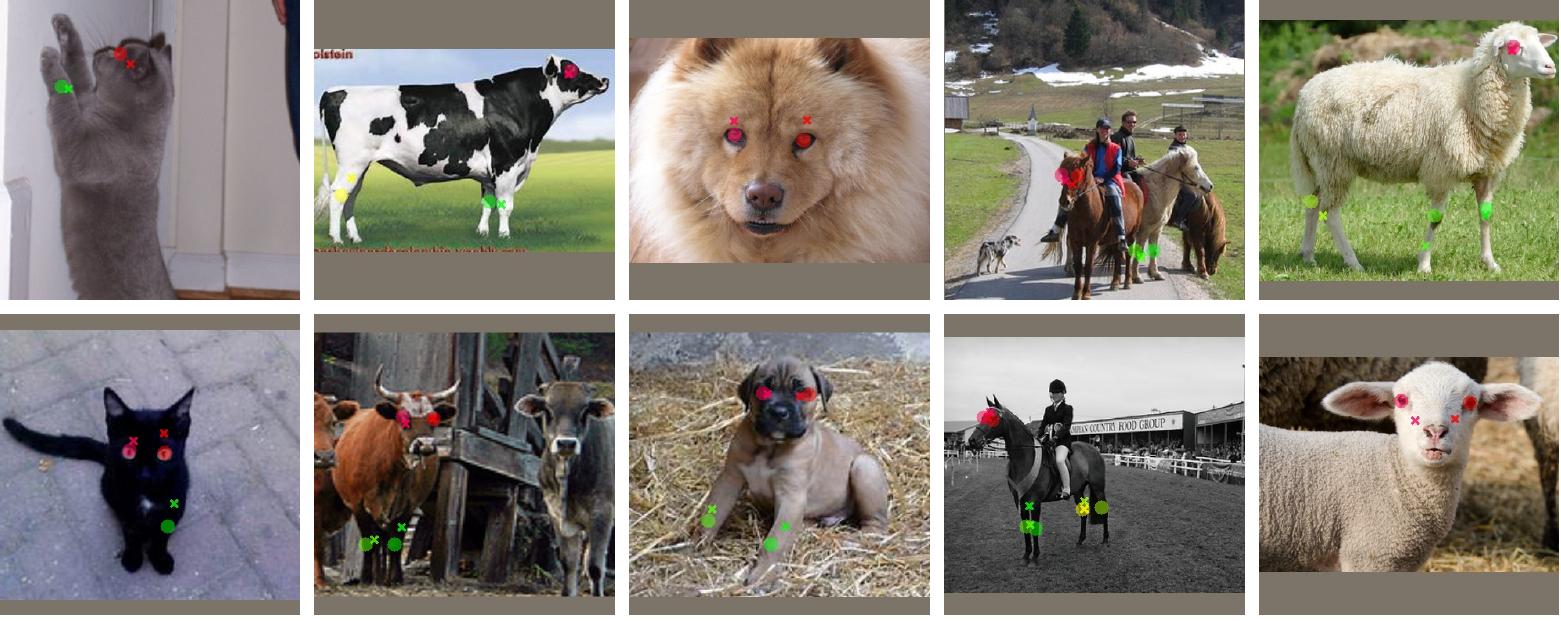}
\caption{
Visualising detection (\ding{54}) and ground-truth (\ding{108}) for novel keypoints for base classes (\textit{top}) and unseen classes (\textit{bottom}).
\vspace{-1.75em}
}\label{fig:visuals}
\vspace{-0.75em}
\end{figure}
A detailed performance comparison of the \textit{naive} baseline B-Vanilla and FSKD~\cite{Lu_2022_CVPR} (repurposed for the cross-modal paradigm) with the proposed framework is presented in \cref{tab:results}. We evaluate in four aforementioned settings for both datasets~\cite{cao2019cross,Ng_2022_CVPR}, and our findings are: \textit{(a)} B-Vanilla performs poorly in the majority of the settings as it lacks robustness in keypoint learning due to the sketch-photo domain gap, and absence of auxiliary keypoints. The significant performance gap comes from the lack of auxiliary keypoints during training, discussed further in \cref{sec:expt_ablation}. \textit{(b)} FSKD~\cite{Lu_2022_CVPR} performs much better than the baseline due to its clever use of localization uncertainty. The huge margin over B-Vanilla can be attributed to uncertainty modeling with main and auxiliary keypoints. However, it falls short of the proposed model due to the sketch-photo domain gap. \textit{(c)} The proposed framework significantly outperforms all its counterparts for all evaluation settings and datasets~\cite{cao2019cross,Ng_2022_CVPR}. This is due to the careful consideration of the sketch-photo domain gap and information maximization across diverse styles. \textit{(d)} All methods have the poorest performance on novel keypoints for unseen classes, which is expected due to the two \textit{levels} of \textit{domain shift} defined by the evaluation protocol. However, our work accounts for the third domain gap between sketches or edgemaps and photos. \textit{(e)} Our method outperforms FSKD~\cite{Lu_2022_CVPR} by $\approx 5$ and $\approx 4$ PCK in Animal Pose~\cite{cao2019cross} and Animal Kingdom~\cite{Ng_2022_CVPR}, respectively, for novel keypoints on unseen classes. A few sample predictions with ground truth from Animal Pose~\cite{cao2019cross} are presented in \cref{fig:visuals}. It also has superior performance over FSKD~\cite{Lu_2022_CVPR}, maintaining a margin of $\approx 4-8$ in all evaluation settings. \textit{(f)} B-Vanilla performs significantly lower for novel keypoints than base keypoints, showing its poor few-shot capability.

\vspace{-0.375em}
\subsection{Generalization on Free-Hand Sketches}
\label{sec:expt_sketch}
\vspace{-0.125em}

\begin{figure}[t]
    \centering
    \includegraphics[width=\linewidth]{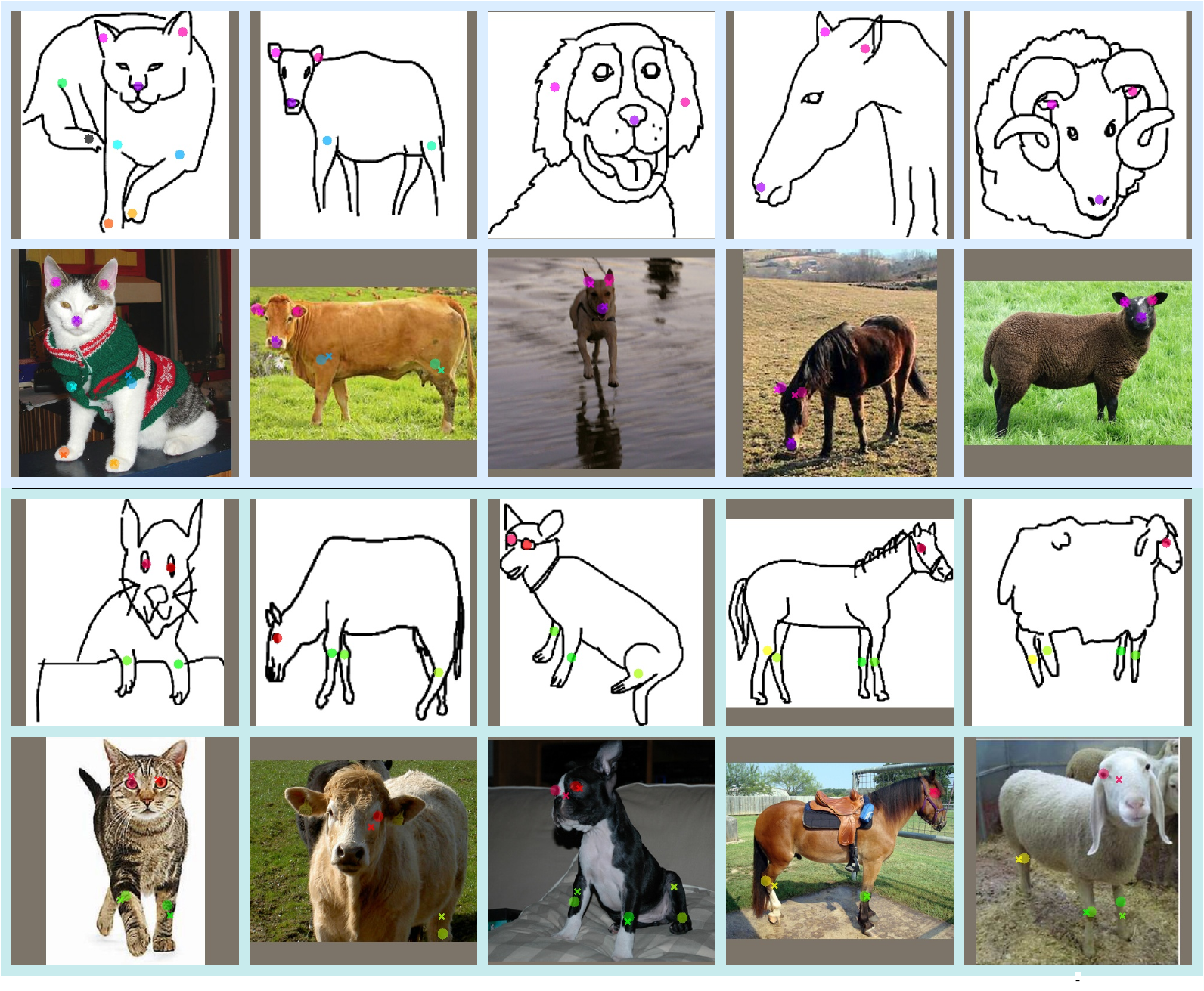}
    \caption{Inference (\ding{54}) with ground-truth (\ding{108}) for base (\textit{top}) and novel (\textit{bottom}) keypoints with annotated sketch prompts.
    \vspace{-2em}
    }
    \label{fig:visuals_sketch}
    \vspace{-1.25em}
\end{figure}

The proposed framework is trained on \textit{synthetic} sketches or edgemaps~\cite{su2021pixel} of the photos in the Animal Pose dataset~\cite{cao2019cross}. Thus, we curate a small subset of \textit{real} sketches of the same classes from the Sketchy database~\cite{sangkloy2016sketchy} to evaluate how well the model generalizes on real free-hand sketches. The evaluation is performed on query photos in the training dataset~\cite{cao2019cross} using a total of 30 real sketches~\cite{sangkloy2016sketchy} with manual keypoint prompts in $K=1$ shot setting. The PCK@0.1 values reported for base and novel keypoints are $42.40\% (\downarrow 0.77)$ and $38.49\% (\downarrow 0.51)$ for unseen classes, and $53.29\% (\downarrow 1.81)$ and $44.99\% (\downarrow 0.15)$ for seen classes respectively, showing the trained model easily adapting to free-hand sketches~\cite{sangkloy2016sketchy}. See \S\hyperlink{./X_suppl.tex}{Suppl.} (\cref{tab:supp_sketch_results}). The inference on a few samples is shown in \cref{fig:visuals_sketch}. The framework, designed for few-shot settings, possesses a high capability for adapting to novel data distributions, which helps it generalize well on real sketches. Sketches are contour-level depiction~\cite{hertzmann2020line} of real photos and are represented closely to edgemaps due to de-stylization (\cref{sec:method_style}), facilitating easy adaptation.
\vspace{-0.5em}
\subsection{Experiments on Multi-modality}
\label{sec:expt_photo}
Considering \textit{novel keypoints on unseen classes}, our framework establishes its superiority with a PCK@0.1 of $39.00\%$ ($\uparrow 5.08$) (\cref{tab:results}) compared to FSKD~\cite{Lu_2022_CVPR} ($33.92\%$). While it is impeccable with sketches, we experiment to see how it fares when being trained with photos instead. FSKD~\cite{Lu_2022_CVPR} being a state-of-the-art on photos with $44.75\%$ PCK@0.1, our model proves to be robust enough to achieve $43.72\%$ ($\downarrow 1.03$) when trained with photos. However, jointly training our model with sketch and photo results in $46.54\%$ outperforming FSKD~\cite{Lu_2022_CVPR} by $1.79$. Evidently, mutual information from both modalities helps in having a more generalized keypoint understanding. Details of the \textit{multi-modal} training and results are in \S\hyperlink{./X_suppl.tex}{Suppl.}
\vspace{-0.25em}
\subsection{Ablation Study}
\label{sec:expt_ablation}
\begin{figure*}[t]
    \begin{floatrow}[4]
    \centering
        \floatbox[\footnotesize]{table}[0.24\textwidth]{%
        \begingroup
        \setlength{\tabcolsep}{1pt}
        \begin{tabular}{lcc}
        \toprule
        \multirow{2}{*}{Method} & $N$ & $N  +T$ \\
        & (only Main)   & (+ Auxiliary)\\
        \midrule
        B-Vanilla       & 17.39 & 29.98 \\
        B-DA            & 18.31 & 31.76 \\
        B-Style         & 18.97 & 32.51 \\
        B-Full          & 19.03 & \textbf{39.00} \\
        \bottomrule
        \end{tabular}
        \endgroup
        \vspace{-1em}
        }{%
        \caption{Performances of incremental baselines.}\label{tab:loss_strip}
        }

        \ffigbox[0.20\textwidth]{%
        \includegraphics[]{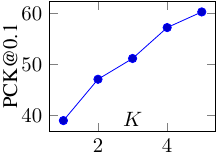}
        \vspace{-2.5em}
        }{%
        \caption{Performance with varying shots.}\label{fig:shots}
        }

        \ffigbox[0.20\textwidth]{%
        \includegraphics[]{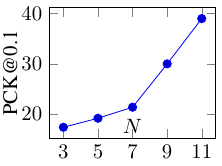}
        \vspace{-2.5em}
        }{%
        \caption{Impact of varying main keypoints.}\label{fig:kp_plot}
        }
        
        \ffigbox[0.20\textwidth]{%
        \includegraphics[]{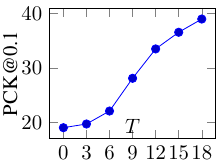}
        \vspace{-2.5em}
        }{%
        \caption{Impact of varying auxiliary keypoints.}\label{fig:kp_plot_aux}
        }
    \end{floatrow}
    \vspace{-1.75em}
\end{figure*}

\vspace{-0.25em}
\keypoint{Number of Shots:}
Typically, increasing the number of shots in a few-shot setting boosts performance~\cite{Lu_2022_CVPR}. We present the empirical data in \cref{fig:shots} that shows performance increase with increasing $K$, although the rate of increment gradually diminishes, denoting \textit{information saturation}.

\keypoint{Significance of Loss Objectives:}
Justifying the independent contribution of the loss objectives, we present a \textit{strip-down-styled} performance analysis of the incremental baselines at $K=1$ shot in \cref{tab:loss_strip}. B-Vanilla is a naive baseline with the poorest performance. B-DA uses domain adaptation~\cite{tanwisuth2021a}, B-Style accounts for style diversity, improving $\approx 1-1.5$ over B-Vanilla. B-Full devises both modules (\cref{sec:method_da,sec:method_style}), has superior performance. Evidently, auxiliary keypoints~\cite{Lu_2022_CVPR} (\cref{sec:method_aux}) significantly boost performance by enhancing the capability of other modules.

\keypoint{Number of Keypoints:}
Increasing the number of keypoints manifests better \textit{keypoint diversity}, resulting in better generalization with improved performance (\cref{fig:kp_plot,fig:kp_plot_aux}). This is also coherent with auxiliary keypoints that, being synthetic in nature, provide robustness to the keypoints learning regime in a semi-supervised setup. Unlike Lu~\etal~\cite{Lu_2022_CVPR}, we augment keypoints for the enhancement of the sub-module performances, resulting in improvements of $\approx 12-20$ for all the baselines (\cref{tab:loss_strip}). Clearly, better-performing baselines improve more than others.

\keypoint{Viability of Edgemaps:}
As we have used edgemaps as synthetic sketch data to train our framework, to justify the choice of edge detectors, \ie~PiDiNet~\cite{su2021pixel}, HED~\cite{xie2015holistically}, and Canny~\cite{canny1986computational}, we experimented with a large number of algorithms, and this particular combination proved to have the best performance. Moreover, our framework is flexible to incorporate superior algorithms from future research.

\keypoint{Further Insights:}
\textit{(a)} Albeit the support keypoint representations $\Phi_{k,n}$ are disentangled to $\delta_{k,n}$ by $Z$, the query keypoints $\hat{\Phi}_{m,n}$ from photos do not have style or sparsity associated and thus do not require de-stylization. Alongside learning robust support keypoint representations $\delta_{k,n}$, this implicitly poses a support-query embedding alignment restricting $\mathcal{L}_{\text{DA}}$ from collapsing.
\textit{(b)} Computation of $\hat{\mu}_n$ for domain adaptation~\cite{tanwisuth2021a} (\cref{sec:method_da}) is detached from the computation graph to prevent a collapse caused by simultaneous gradients through the query and support prototypes.
\textit{(c)} The architecture of the de-stylization network (\cref{sec:method_style}) $Z$ is non-trivial as complex models lead to unstable training, and simpler ones are incapable of style disentanglement. (See \S\hyperlink{./X_suppl.tex}{Suppl.})
\textit{(d)} The additional edgemaps $x^{\texttt{S}_1}_i$ and $x^{\texttt{S}_2}_i$ are only used in de-stylization loss $\mathcal{L}_{\text{style}}$, but not in localization loss $\mathcal{L}_{\text{KP}}$ due to significant performance degradation.
\vspace{-0.5em}
\section{Conclusion}
\label{sec:conclusion}
\vspace{-0.25em}
The few-shot approach to keypoint localization being fairly new in the research community, we explore the scope of \textit{cross-modal} few-shot keypoint detection, adapting quickly to novel keypoints as well as unknown image classes. To this end, we propose a framework to detect novel keypoints on query photos from a few annotated sketches, achieving a \textit{source-free} setup in this paradigm. Particularly, we carefully address the sketch-photo domain gap using domain adaptation, besides catering to the style and abstraction diversity of sketches. Despite being trained on synthetic sketch-like data, our method proves to be adapting well for free-hand sketches. Furthermore, we show that multi-modal training helps generalize better on query photos. To the best of our knowledge, this is amongst the first of the works on cross-modal few-shot keypoint learning, and we hope our work inspires future research in this context.

{
    \small
    \bibliographystyle{ieeenat_fullname}
    \bibliography{main}
}

\clearpage
\maketitlesupplementary

\section{Insights on Framework Design}
\label{sec:supp_arch}
\subsection{Transport Loss for Keypoint Localization}
\label{sec:supp_uda4kp}
\keypoint{Background:}
Prototypical domain adaptation~\cite{tanwisuth2021a} is particularly curated for classification that differs from keypoint detection~\cite{Lu_2022_CVPR} significantly at the task level. Thus, to adapt the same for the sketch-photo domain gap, we needed to analyze every piece of the mechanics it is principled on. The bi-directional transport loss~\cite{tanwisuth2021a} precisely employs two transport objectives: one to pull the samples in the target domain towards the source domain prototype, and the other to pull the prototype towards respective targets. Taking the unsupervised setting into account, the source prototype is considered to be an overall class prototype that encompasses both the source and target domains.

It is worth noting that the prototype formulation in prototypical domain adaptation~\cite{tanwisuth2021a} is particularly dependent on the \textit{weights of the last linear layer} responsible for performing classification, while the conventional prototype formulation~\cite{snell2017prototypical} uses an \textit{mean of the vector representations} in the latent space. Although in both cases, prototypes are essentially built in the same latent embedding, the metric relation between the prototype and the source embedding is completely disparate. Likewise, the original prototypical network~\cite{snell2017prototypical} can be used with any distance metric, while the former~\cite{tanwisuth2021a} is restricted to element-wise multiplication between source embeddings and prototypes or weights.

\keypoint{Dissecting Bidirectional Transport Loss:}
Considering $\vartheta_j(\alpha_j) = \frac{(\alpha_j)}{\sum(\alpha_j)}$ as a normalized weighing function, the bidirectional transport loss~\cite{tanwisuth2021a} is given as:
\vspace{-0.5em}
{
\small
\begin{equation}
\label{equ:supp_uda_loss}
\begin{aligned}
    \mathcal{L}_{t \rightarrow \mu} + &\mathcal{L}_{\mu \rightarrow t} =\\
    &\mathbb{E}[\sum_n \mathbf{C}(\mu_n, \hat{\Phi}_{m,n})\cdot\vartheta_n(p(\mu_n)\exp(\mu_n\hat{\Phi}_{m,n}^{\mathbf{T}}))] +\\
    &\mathbb{E}[\sum_n p(\mu_n) \sum_m \mathbf{C}(\mu_n, \hat{\Phi}_{m,n}) \vartheta_m(\exp(\mu_n\hat{\Phi}_{m,n}^{\mathbf{T}}))]
\end{aligned}
\vspace{-0.25em}
\end{equation}
}
Here, the terms $\mathcal{L}_{t \rightarrow \mu}$ and $\mathcal{L}_{\mu \rightarrow t}$ signify two directions, target domain to prototypes and prototypes to target domain, respectively, of the bi-directional aspect of the loss function in \cref{equ:supp_uda_loss}. The term $\mathbf{C}(\mu_n, \hat{\Phi}_{m,n})$ refers to a direct \textit{point-to-point} cost between a prototype and the corresponding target embeddings and implements a \textit{cosine distance} between prototype $\mu_n$ and respective $m$th target embedding $\hat{\Phi}_{m,n}$ for any class $n$. Also, the expansion of $\mathcal{L}_{\mu \rightarrow t}$ term uses the normalized weighing function  $\vartheta$ to mimic \texttt{softmax} operation implicitly, and the expansion of $\mathcal{L}_{t \rightarrow \mu}$ uses the \texttt{softmax} function and directly provides a discriminative entropy minimization linking to the classification task regime. Moreover, this is especially aligned with the corresponding prototype formulation with weights of the last layer.

As the keypoint detection does not need to predict discriminative probabilities between different classes or keypoints, we remove the weighing function $\vartheta$ and \cref{equ:supp_uda_loss} reduces to:
{\small
\vspace{-1em}
\begin{equation}
\label{equ:supp_uda_loss_reduce}
\begin{aligned}
    \mathcal{L}_{t \rightarrow \mu} + \mathcal{L}_{\mu \rightarrow t} = 
    &\mathbb{E}[\sum_n \mathbf{C}(\mu_n, \hat{\Phi}_{m,n}) p(\hat{\mu}_n)\exp(\mu_n\hat{\Phi}_{m,n}^{\mathbf{T}})] +\\
    &\mathbb{E}[\sum_n p(\hat{\mu}_n) \sum_m \mathbf{C}(\mu_n, \hat{\Phi}_{m,n}) \exp(\mu_n\hat{\Phi}_{m,n}^{\mathbf{T}})]
\end{aligned}
\vspace{-0.25em}
\end{equation}
}
In \cref{equ:supp_uda_loss_reduce}, the term $\exp(\mu_n\hat{\Phi}_{m,n}^{\mathbf{T}})$ is essentially a \textit{similarity measure} that closely resembles the cosine similarity. Taking $\mathbf{Sim}$ as a similarity function, \cref{equ:supp_uda_loss_reduce} can be further expressed as \cref{equ:supp_uda_loss_reduce2}, identifying the key factors responsible for the domain adaptation in the transport loss.
{\small
\vspace{-0.75em}
\begin{equation}
\label{equ:supp_uda_loss_reduce2}
\begin{aligned}
    \mathcal{L}_{t \rightarrow \mu} + \mathcal{L}_{\mu \rightarrow t} =
    &\mathbb{E}[\sum_n \mathbf{C}(\mu_n, \hat{\Phi}_{m,n}) p(\hat{\mu}_n) \mathbf{Sim}(\mu_n, \hat{\Phi}_{m,n})]\\
    +&\mathbb{E}[\sum_n p(\hat{\mu}_n) \sum_m \mathbf{C}(\mu_n, \hat{\Phi}_{m,n}) \mathbf{Sim}(\mu_n, \hat{\Phi}_{m,n})]
\end{aligned}
\vspace{-0.5em}
\end{equation}
}

\keypoint{Adaptation to Keypoints Learning Paradigm:}
From \cref{equ:supp_uda_loss_reduce2}, it is clear that both the loss terms, $\mathcal{L}_{t \rightarrow \mu}$ and $\mathcal{L}_{\mu \rightarrow t}$ are reduced to the similar form consisting of target prototype $p(\hat{\mu}_n)$, a point-to-point cost $\mathbf{C}(\mu_n, \hat{\Phi}_{m,n})$ and a similarity score $\mathbf{Sim}(\mu_n, \hat{\Phi}_{m,n})$. Thus, having a similar form of both transport losses~\cite{tanwisuth2021a}, the domain adaptation loss $\mathcal{L}_{\text{DA}}$ in \cref{equ:loss_da} takes a unified form of the transport losses. Considering the aforementioned inherent differences at the task level, $\mathbf{C}(\mu_n, \hat{\Phi}_{m,n})$ and $\mathbf{Sim}(\mu_n, \hat{\Phi}_{m,n})$ are realized by $l_2$ distances and a derived similarity score for the keypoint learning task as mentioned in details in \cref{sec:method_da}.

It is also to be noticed that $\hat{\mu}_n$ is used in \cref{equ:supp_uda_loss_reduce,equ:supp_uda_loss_reduce2} to represent the target prototype. Due to the absence of class information of the target domain, Tanwisuth~\etal~\cite{tanwisuth2021a} uses a single prototype system for source and target domains. We consider the gradual movement of prototypes from the source to the target domain, and thus we replace the same with a prototype calculation for our target domain,~\ie~query keypoints using \cref{equ:proto_prob}. The term $p(\mu_n)$ essentially represents the \textit{probability} of the prototypes given all the target samples. Given the task-level setting of unsupervised classification, $p(\mu_n)$ is interpreted as class proportion in the original bi-directional transport loss~\cite{tanwisuth2021a} and is iteratively updated starting from a uniform class distribution. On the contrary, keypoints learning replaces it by an equivalent term $p(\hat{\mu}_n)$ which refers to $l_2$ distance-based probability for the prototypes (\cref{equ:proto_prob}) and it could be calculated dynamically using the keypoint-level class information of the query photos, turning the problem setup to a supervised one and dismissing the necessity of iterative updates of the prototype likelihood.
\subsection{Design Specifications of Descriptor Network}
\label{sec:supp_descriptor_specs}
The descriptor network~\cite{Lu_2022_CVPR} $D$ is particularly employed to \textit{refine and encode} the features at the local scale from the correlated query feature maps $\mathcal{A}_{m,n}$ having dense encoded features pertaining to both the query feature map $f_m$ and the support prototype $\mu_n$, to a descriptor $\Psi_{m,n}$ so that it contains necessary positional information to localize the relevant keypoint $n$ in query photo $x_m$.

The architecture of the descriptor network $D$ is taken from FSKD~\cite{Lu_2022_CVPR}, and the design specification is kept the same as well. The network $D$ consists of three consecutive convolution layers with kernel size $3\times3$, a stride of 2, a padding of 1, and \texttt{ReLU} as activation. The input channels for the first convolution layer are $c=2048$, the output channels of the last layer are 1024, and all the intermediate input or output channels are 512. Considering the input size of $x_i$ being $384\times384$ we have correlated query features $f_m$ of size $\mathbb{R}^{2048\times12\times12}$ as input of descriptor network $D$ which results in tensors of $\mathbb{R}^{512\times6\times6}$, $\mathbb{R}^{512\times3\times3}$ and $\mathbb{R}^{1024\times2\times2}$ as the consecutive outputs of convolution layers. Thus, the final output descriptor $\hat{\Psi}_{m,n}$ of dimension $d=4096$ is formed by flattening the last output from $D$.
\subsection{Architecture Choice for De-stylization Network}
\label{sec:supp_destyle}
\keypoint{Background:}
The de-stylization network $Z$ as per the architecture given in \cref{fig:style} is designed after the multi-scale channel attention module~\cite{dai2021attentional} to fuse the keypoint level local information with the global context of sparsity and style present in a sketch or edgemap. While multi-scale channel attention~\cite{dai2021attentional} is specifically curated for convolutional feature maps, our design needs to deal with keypoint embeddings of $\mathbb{R}^c$, keeping a similar notion of context fusion~\cite{dai2021attentional}. The original design~\cite{dai2021attentional} uses two parallel branches on convolutional feature maps, one with the input feature map as it is, and the other using a globally pooled vector from the input convolutional feature map, encoded with separate learnable convolution layers and both the branches are aggregated using an element-wise addition for fusing the global context into feature maps. The resulting convolutional map is then passed through \texttt{sigmoid} activation to adjust the weight to which the fused context should affect the original input map while it is multiplied with the fused feature map.

\keypoint{Proposed Design:}
Designing our de-stylization network $Z$ (see \cref{fig:style}), we need it to cater to the keypoint embeddings of $\mathbb{R}^c$ with dense local features, and thus they are fused with globally pooled vectors from the corresponding feature map $f_k$. The context fusion for any keypoint $n$ in $x_k$ is achieved by the concatenation of the extracted keypoint embedding $\Phi_{k,n}$ and the corresponding global pooled features from $f_k$, followed by two linear layers with a \texttt{ReLU} activation in between. This, in particular, is used as a context at both \textit{local} and \textit{global} scales and is added element-wise to $\Phi_{k,n}$ and a \texttt{sigmoid} activation, along with two more linear layers with a \texttt{ReLU} activations in between help in learning keypoint embeddings $\delta_{k,n}$ from the dense fused features at local and global scales.
\section{Design Analysis of De-stylization Network}
\label{sec:supp_style_justify}
\begin{table}[t]
    \caption{A comparison of performance for different architecture designs of de-stylization network $Z$ along with the usage of local and global contexts. The PCK@0.1 is measured on the Animal Pose dataset~\cite{cao2019cross} for novel keypoints on unknown classes.
    \vspace{-2em}
    }
    \label{tab:supp_destyle_compare}
    \centering
    \footnotesize
    \begingroup
    \setlength{\tabcolsep}{5pt}
    \begin{tabular}{lccc}
    \toprule
    Architecture of $Z$   & Local Context & Global Context & PCK@0.1 \\
    \midrule
    B-DA (No $\mathcal{L}_{\text{style}}$) & \cmark        & \xmark         & 31.76 \\
    None (Identity)                        & \cmark        & \xmark         & 36.84 \\
    MLP                                    & \cmark        & \xmark         & 37.78 \\
    MLP (Concatenated)                     & \cmark        & \cmark         & 38.11 \\
    \textbf{Proposed}                      & \cmark        & \cmark         & \textbf{39.00} \\
    \bottomrule
    \end{tabular}
    \endgroup
\vspace{-1.25em}
\end{table}
The de-stylization network $Z$ disentangles style and sparsity from the keypoint embeddings using \textit{attentional} global and local \textit{context fusion}~\cite{dai2021attentional}. However, to understand the role of global context realized by the global pooled vector from the support feature $f_k$ and also to justify our choice of architecture, we designed a few different architectures of $Z$ and rigorously experimented and evaluated for novel keypoints on unseen classes (\cref{tab:supp_destyle_compare}).

\textit{(a)} The baseline B-DA is used as a control measure as there is no $Z$ or $\mathcal{L}_{\text{style}}$ involved. \textit{(b)} Using an identity function as $Z$, we essentially ensure $\Phi_{k,n} = \delta_{k,n}$. However, the framework still tries to learn style-agnostic keypoint embeddings using the style loss $\mathcal{L}_{\text{style}}$ with additional edgemaps, which have a significant performance upgradation ($\uparrow 5.08$) over B-DA. This proves that the loss $\mathcal{L}_{\text{style}}$ contributes significantly and thus, the assumption regarding existing style diversity in the synthetic sketches or edgemaps is validated. \textit{(c)} Using a multi-layer perceptron (MLP) as an alternative architecture of $Z$ is taken into consideration. While the MLP takes only extracted keypoint embedding $\Phi_{k,n}$ as input, it has an improvement of 0.94 over the case where $\Phi_{k,n} = \delta_{k,n}$. This signifies the learning of better keypoints with $Z$ while preserving \textit{alignment} between the support keypoint and the query. \textit{(d)} Using the MLP designed to take the dense features, formulated by concatenation of global pooled features and the extracted keypoint embeddings, the performance gets a further boost of 0.33, which proves that global context has a significant role in disentanglement and can encode the style and sparsity information. \textit{(e)} The proposed design architecture has the best performance with an improvement of 0.89 over MLP with a global context aggregation. \textit{(f)} Apart from the architectures presented in the \cref{tab:supp_destyle_compare}, we also experimented with several other attention modules. However, all such models had \textit{intractable} training with exploding gradients and loss values greater than $10^8$ times the usual.
\vspace{-0.25em}
\section{Choosing Edge Detection Algorithms}
\label{sec:supp_edge}
The choice of edge detection algorithm is particularly crucial for our problem, as it directly connects with the training data for the proposed framework. As we treat edgemaps as \textit{synthetic} sketch data, we performed in-depth experimentation with the different edge detection algorithms as mentioned in \cref{sec:expt_ablation}. Precisely, we include modern and popular edge detector algorithms, including Im2pencil~\cite{li2019im2pencil}, DeXiNed~\cite{xsoria2020dexined}, Photo-Sketch~\cite{li2019photo}, HED~\cite{xie2015holistically}, Canny~\cite{canny1986computational}, PiDiNet~\cite{su2021pixel} in multiple combinations. The top 5 performing combinations on both datasets~\cite{cao2019cross,Ng_2022_CVPR} for the most challenging evaluation setting (\textit{novel} keypoints on \textit{unseen} classes) are given in \cref{tab:supp_edgemaps}. While the proposed combination performs the best, the presented data shows that the performance does not vary by more than 3 PCK@0.1 for both datasets~\cite{cao2019cross,Ng_2022_CVPR}, essentially indicating that the proposed framework is robust enough to accommodate different levels of style and sparsity in sketch data. Moreover, we have experimented with real photos in the support set instead of sketches or edgemaps, achieving a similar accuracy of 43.72\% (only a gain of $\uparrow 4.72$ over sketches) on the Animal Pose dataset~\cite{cao2019cross} for novel keypoints on query photos of unseen classes. See \cref{sec:expt_photo,sec:supp_multimodal} for detailed results and analysis. Thus, it could be argued that the proposed method is \textit{robust} and capable enough to \textit{encode keypoint-level information} across diverse styles of sketches as well as photos. While we understand the capability of encoding synthetic sketches, the following section (\cref{sec:supp_freehand}) illustrates the practicality aspect of using synthetic sketch data or edgemaps.
\begin{table}[t]
    \footnotesize
    \centering
    \caption{Performance of the proposed framework with various edge detection algorithms used for generating synthetic sketch data for training. Performances (PCK@0.1) are measured on both datasets~\cite{cao2019cross,Ng_2022_CVPR} for novel keypoints on unseen species.
    \vspace{-1em}
    }
    \label{tab:supp_edgemaps}
    \setlength{\tabcolsep}{3pt}
    \begin{tabular}{lcc}
    \toprule
    Edge Detector & Animal Pose~\cite{cao2019cross} & Animal Kingdom~\cite{Ng_2022_CVPR}\\
    \midrule
    \cite{li2019im2pencil} + \cite{xsoria2020dexined} + \cite{xie2015holistically}                      & 36.87 & 12.46 \\
    \cite{xie2015holistically} + \cite{li2019photo} + \cite{canny1986computational}                     & 37.33 & 12.98 \\
    \cite{canny1986computational} + \cite{li2019im2pencil} + \cite{xsoria2020dexined}                   & 38.38 & 13.45 \\
    \cite{su2021pixel} + \cite{li2019im2pencil} + \cite{li2019photo}                                    & 38.61 & 13.87 \\
    \textbf{Ours} (\cite{su2021pixel} + \cite{xie2015holistically} + \cite{canny1986computational})     & \textbf{39.00} & \textbf{14.42} \\
    \bottomrule
    \end{tabular}
\vspace{-1.25em}
\end{table}
\section{Empirical Study with Free-Hand Sketches}
\label{sec:supp_freehand}
\begin{figure*}[t!]
    \centering
    \includegraphics[width=\textwidth]{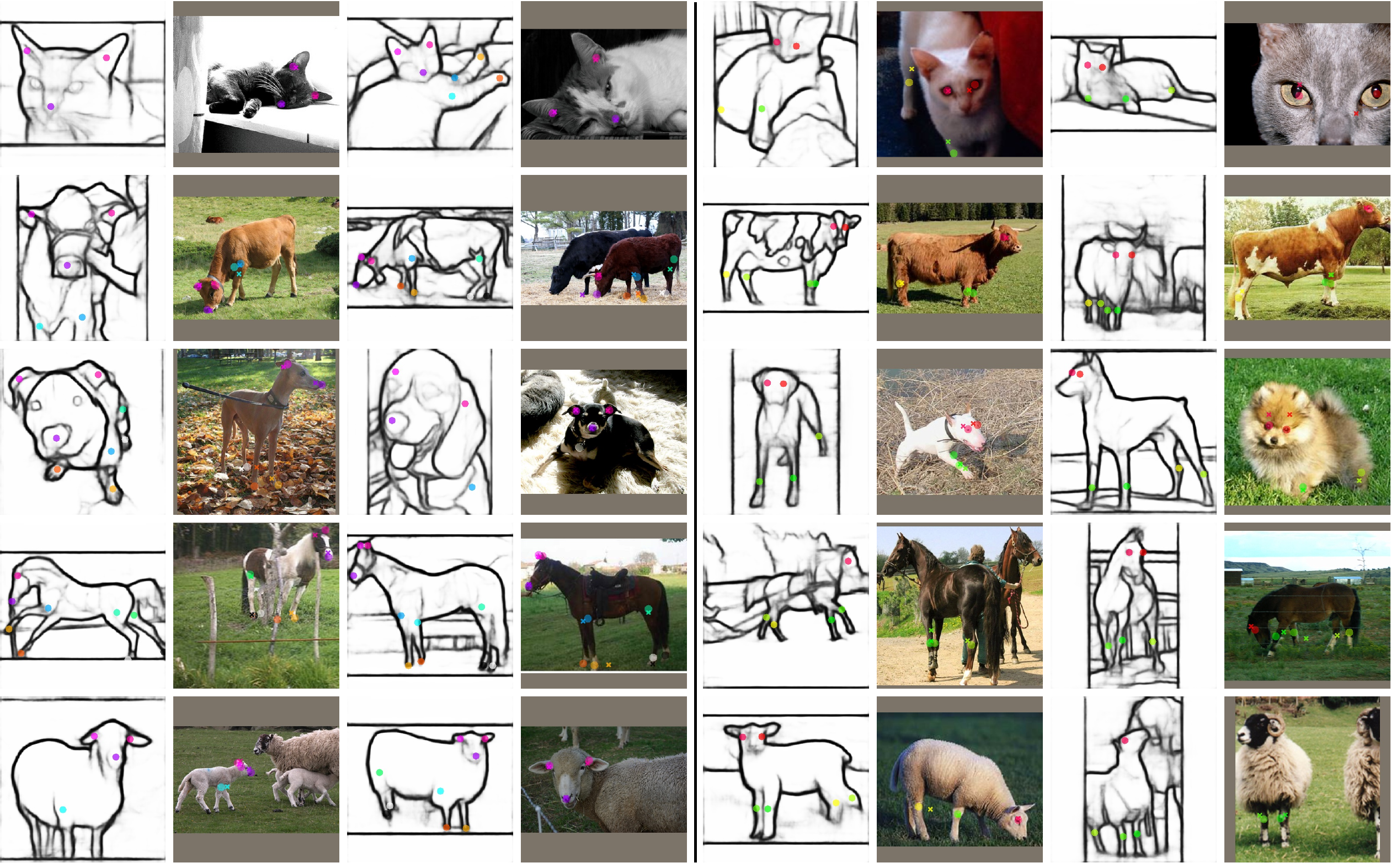}
    \vspace{-2em}
    \caption{Visualizations of sample detection (\ding{54}) along with ground-truth (\ding{108}) for base (left) and novel (right) keypoints on query photos~\cite{cao2019cross} using support edgemaps~\cite{su2021pixel}.
    \vspace{-1em}
    }
    \label{fig:supp_visuals_edge}
\end{figure*}

\begin{figure*}[t!]
    \centering
    \includegraphics[width=\textwidth]{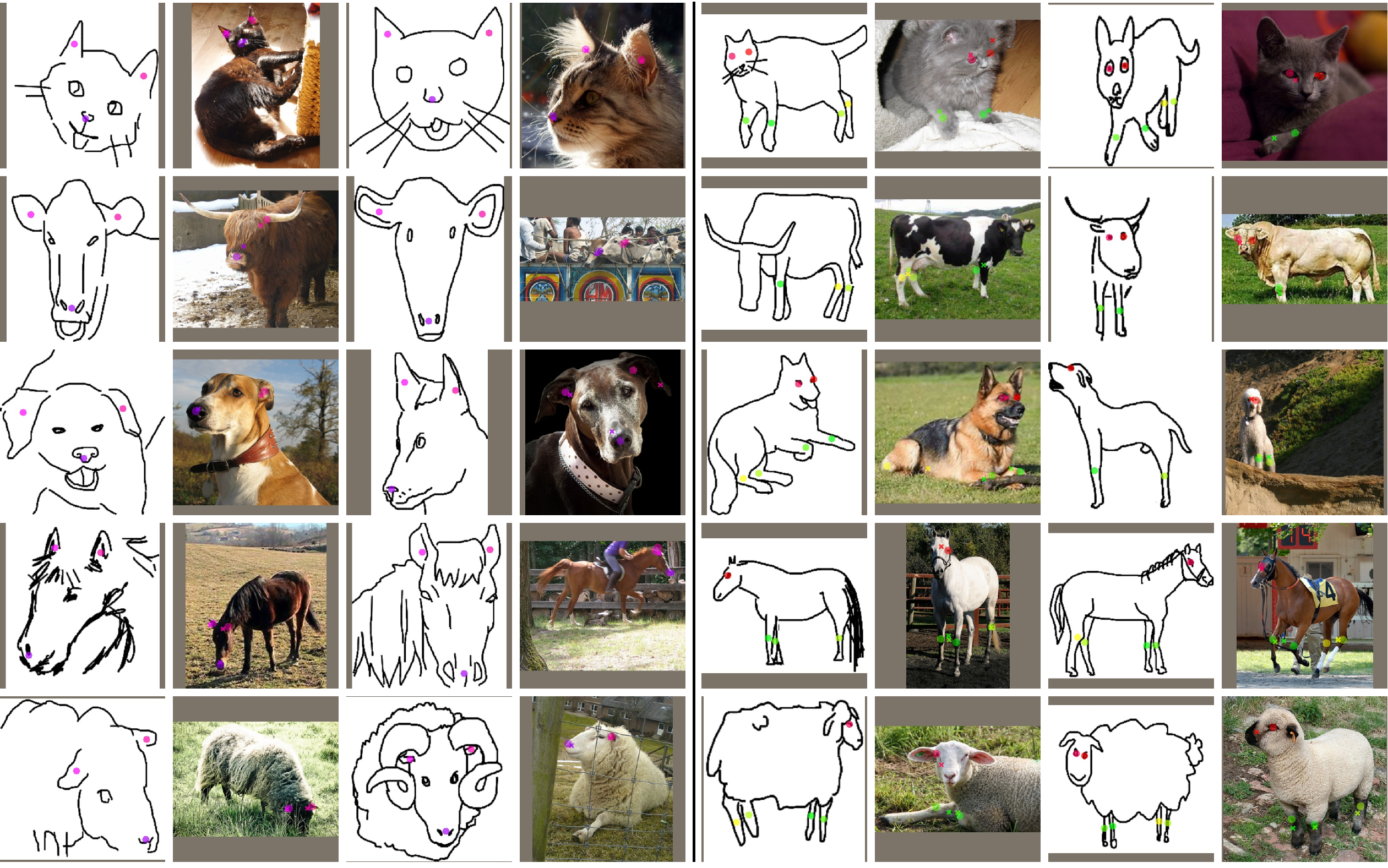}
    \vspace{-2em}
    \caption{Visualizations of sample detection (\ding{54}) along with ground-truth (\ding{108}) for base (left) and novel (right) keypoints on query photos~\cite{cao2019cross} using real support sketches~\cite{sangkloy2016sketchy} with manual annotation prompt.
    \vspace{-1em}
    }
    \label{fig:supp_visuals_sketch}
\end{figure*}

The proposed framework is entirely trained with edgemaps or synthetic sketches as given in \cref{sec:method,sec:expt}, and a few sample visualizations of support edgemap~\cite{su2021pixel} and detection with ground-truth on query photos~\cite{cao2019cross} are given in \cref{fig:supp_visuals_edge}. Thus, we perform extended experimentations with pre-trained models as described in \cref{sec:expt_sketch} along with a sample depiction of base and novel keypoints on support sketches~\cite{sangkloy2016sketchy} and query photos~\cite{cao2019cross} in \cref{fig:visuals_sketch}.
From the data presented in \cref{tab:supp_sketch_results}, it is evident that our framework has \textit{adequate} few-shot capability for generalizing across real support sketches, as PCK with $\tau=0.1$ for real sketches is within a range of 5 from the extensive evaluation results on edgemaps. More visualizations for base and novel keypoints are in \cref{fig:supp_visuals_sketch}.

\begin{table}[t]
    \footnotesize
    \centering
    \begingroup
    \setlength{\tabcolsep}{2pt}
    \begin{tabular}{lllcccccg}
    \toprule
    \multirow{2}{*}{Class}  & \multirow{2}{*}{Keypoints} & \multirow{2}{*}{Support} & \multicolumn{6}{c}{PCK@0.1} \\
    \cmidrule{4-9}
                            &                            &                  & Cat   & Cow   & Dog   & Horse & Sheep & \textbf{Mean}\\
    \midrule
    \multirow{4}{*}{Seen}   & \multirow{2}{*}{Base}      & Edgemap          & 67.34 & 49.89 & 56.28 & 56.35 & 45.65 & 55.10 \\
                            &                            & Sketch           & 66.69 & 45.79 & 55.43 & 56.13 & 43.40 & 53.29 \\
    \cmidrule{2-9}
                            & \multirow{2}{*}{Novel}     & Edgemap          & 55.69 & 43.09 & 46.58 & 43.94 & 36.39 & 45.14 \\
                            &                            & Sketch           & 55.45 & 42.96 & 46.35 & 43.88 & 36.31 & 44.99 \\
    \midrule
    \multirow{4}{*}{Unseen} & \multirow{2}{*}{Base}      & Edgemap          & 47.36 & 42.97 & 38.30 & 46.17 & 41.03 & 43.17 \\
                            &                            & Sketch           & 45.90 & 42.47 & 37.82 & 45.36 & 40.45 & 42.40 \\
    \cmidrule{2-9}
                            & \multirow{2}{*}{Novel}     & Edgemap          & 44.42 & 40.13 & 36.91 & 37.77 & 35.77 & 39.00 \\
                            &                            & Sketch           & 43.79 & 39.91 & 36.17 & 37.56 & 35.02 & 38.49 \\
    \bottomrule
    \end{tabular}
    \endgroup
    \vspace{-0.75em}
    \caption{A quantitative comparison of the proposed method on query photos~\cite{cao2019cross} using edgemaps and real free-hand sketches~\cite{sangkloy2016sketchy} with $K=1$ for all evaluation settings.
    \vspace{-1.5em}
    }
    \label{tab:supp_sketch_results}
\end{table}

\section{Experiments with Support Photos}
\label{sec:supp_multimodal}
\begin{table}[b]
\vspace{-1.25em}
    \footnotesize
    \centering
    \begingroup
    \setlength{\tabcolsep}{1.8pt}
    \begin{tabular}{lllcccccg}
    \toprule
    \multirow{2}{*}{Class}  & \multirow{2}{*}{Keypoints} & \multirow{2}{*}{Method}  & \multicolumn{6}{c}{PCK@0.1} \\
    \cmidrule{4-9}
                            &                            &                          & Cat   & Cow   & Dog   & Horse & Sheep & \textbf{Mean}\\
    \midrule
    \multirow{6}{*}{Seen}   & \multirow{3}{*}{Base}      & FSKD~\cite{Lu_2022_CVPR} & 68.66 & 52.70 & 59.24 & 58.53 & 45.04 & 56.83 \\
                            &                            & Ours                     & 66.97 & 51.38 & 57.72 & 57.31 & 43.81 & 55.44 \\
                            &                            & Ours (MM)                & \textbf{80.16} & \textbf{61.34} & \textbf{73.70} & \textbf{67.44} & \textbf{57.85} & \textbf{68.10} \\
    \cmidrule{2-9}
                            & \multirow{3}{*}{Novel}     & FSKD~\cite{Lu_2022_CVPR} & 60.84 & 47.78 & 53.44 & 49.21 & 38.47 & 49.95 \\
                            &                            & Ours                     & 59.17 & 46.49 & 51.89 & 47.93 & 37.65 & 48.63 \\
                            &                            & Ours (MM)                & \textbf{67.51} & \textbf{49.92} & \textbf{59.05} & \textbf{53.06} & \textbf{43.45} & \textbf{54.60} \\
    \midrule
    \multirow{6}{*}{Unseen}& \multirow{3}{*}{Base}       & FSKD~\cite{Lu_2022_CVPR} & 56.38 & 48.24 & 51.29 & 49.77 & 43.95 & 49.93 \\
                            &                            & Ours                     & 55.67 & 46.94 & 50.47 & 48.21 & 42.88 & 48.83 \\
                            &                            & Ours (MM)                & \textbf{57.68} & \textbf{52.06} & \textbf{51.75} & \textbf{52.27} & \textbf{47.74} & \textbf{52.30} \\
    \cmidrule{2-9}
                            & \multirow{3}{*}{Novel}     & FSKD~\cite{Lu_2022_CVPR} & 52.36 & 44.07 & 47.94 & 42.77 & 36.60 & 44.75 \\
                            &                            & Ours                     & 50.88 & 43.34 & 46.67 & 42.52 & 35.19 & 43.72 \\
                            &                            & Ours (MM)                & \textbf{54.61} & \textbf{45.92} & \textbf{48.02} & \textbf{43.86} & \textbf{40.31} & \textbf{46.54} \\
    \bottomrule
    \end{tabular}
    \endgroup
    \caption{A quantitative comparison of the proposed method on query photos~\cite{cao2019cross} using photo only and both edgemap and photos (MM) with the FSKD~\cite{Lu_2022_CVPR} in $K=1$ for all evaluation settings.
    }
    \label{tab:supp_photo_results}
\vspace{-0.75em}
\end{table}

Apart from using sketches or edgemaps as support, we also experiment with photos as support, solving the simple few-shot keypoint detection problem Lu~\etal~\cite{Lu_2022_CVPR} solves. This experimentation was conducted to prove the robustness of the proposed work. While photos do not have any style or abstraction diversity, we first experiment with only photos in a setting similar to the original work of FSKD~\cite{Lu_2022_CVPR}. In this setting, we completely turn off the de-stylization loss $\mathcal{L}_{\text{style}}$ (\cref{sec:method_style}) due to lack of additional support inputs. However, the de-stylization network $Z$ being an integral part of the architecture keeps aiding the learning of deeper features. Empirically, although our method goes close to the state-of-the-art FSKD~\cite{Lu_2022_CVPR}, it fails to outperform. Next, we carefully devise the experimentation strategy using edgemaps~\cite{su2021pixel,xie2015holistically} as additional sketches turning on the de-stylization loss $\mathcal{L}_{\text{style}}$ (with reduced weight of $\lambda_{\text{style}}=10^{-8}$) to learn the keypoint representation with \textit{mutual information} from both photos and edgemaps. In this scenario, our method seems to outperform FSKD~\cite{Lu_2022_CVPR} by a significant margin ($\approx 2 - 11$) in all evaluation settings, proving the superiority of the multi-modal paradigm~\cite{lu2024openkd}.


\section{Style Diversity in Real Sketches}
\label{sec:supp_human}
We simulate user sketch styles~\cite{sain2021stylemeup} with different edgemaps as the de-stylization network (\cref{sec:method_style}) disentangles the style-invariant features. In order to understand its style-invariance and generalization capability to real sketches, we perform a \textit{human study}, where each of the $20$ participants was asked to draw $10$ sketches, totalling $200$ sketches. The participants were asked to rate every keypoint predicted on the query photos by the models in question, on a scale of $1$$\rightarrow$$5$ (bad$\rightarrow$excellent) based on their \textit{opinion} of how closely it matched their expectation. The proposed method achieves an average score of $4.42$, compared to $2.91$ of the B-Vanilla and $3.38$ of the FSKD~\cite{Lu_2022_CVPR}, underpinning the \textit{generalizability} and \textit{user-style independence} of our work. While we have used a limited number of real sketches from the Sketchy Extended~\cite{sangkloy2016sketchy} database, this study further proves the \textit{practicality} of our framework.

\section{Challenges with Additional Modalities}
\label{sec:supp_modalities}
Our few-shot framework is particularly curated for sketch-photo cross-modal learning, and is vastly different from traditional sketch research~\cite{bhunia2020sketch,bhunia2021more,chowdhury2022partially}, as careful addressing of the domain shift is well-observed in sketch-photo cross-modal literature~\cite{lu2023learning,richardson2021encoding,li2023photo}, depending on tasks. The major sketch applications like sketch-based image retrieval~\cite{sain2023clip,sain2023exploiting,yu2016sketch} conventionally use a \textit{joint representation} space due to the availability of instance-level sketch-photo pairs. However, \textit{without} overlap of support and query sets by definition, the unavailability of such \textit{instance-level sketch-photo pairs} necessitates the need for explicit keypoint-level domain adaptation (see \cref{sec:method_da}) using a transport loss~\cite{tanwisuth2021a} on the de-stylized keypoint representation, accounting for the unique sparse nature of sketches.

This sketch-photo cross-modal domain adaptation becomes challenging when extended to other modalities. OpenKD~\cite{lu2024openkd} utilizes \textit{text} as an additional guidance, along with annotated support photos, resulting in a multi-modal setup. In the context of sketch, we have experimented with a similar multi-modal setting with sketch and photo in \cref{sec:expt_photo,sec:supp_multimodal}. Taking the inspiration from OpenKD~\cite{lu2024openkd}, we attempted \textit{text-to-photo} cross-modal keypoint learning in our framework using an off-the-shelf frozen CLIP textual encoder~\cite{radford2021learning} to obtain textual keypoint embeddings that replace support prototypes. The rest of the framework follows B-Vanilla (\cref{sec:method_framework}), using the feature modulation $\mathcal{M}$ to correlate the textual support prototypes with query features from the image encoder $F$, followed by the descriptor network $D$, and a GBL module for localization. This particular experiment achieves $20.13\%$ ($\uparrow 2.74$ over B-Vanilla, $\downarrow 18.87$ below proposed) for the novel keypoints on unseen classes in the Animal Pose~\cite{cao2019cross} dataset. This poor performance is expected due to certain factors. Firstly, OpenKD~\cite{lu2024openkd} uses the annotated RGB photos as support along with text in a multi-modal setup. Incorporating it into a \textit{source-free} paradigm is challenging, as the \textit{text-photo joint keypoint representation} becomes unavailable due to the absence of such pairs in the support set. Secondly, our framework is designed to handle \textit{image-like} data,~\eg~sketch and photo, but it is not equipped to handle textual data, and needs explicit text-photo cross-modal design.

\section{Additional Comparisons}
\label{sec:supp_add_sota}
Although few-shot keypoint learning~\cite{sun2024uniap,ge2021metacloth,lu2023saliency} has been around for some time, only a few state-of-the-art methods are suitable for comparison to the proposed method. Our framework follows FSKD~\cite{Lu_2022_CVPR} closely, and we compare them in \cref{tab:results}. Additionally, we adapt ProbIntr~\cite{novotny2018self} to a few-shot framework. We also consider GeometryKP~\cite{he2023few} and OpenKD~\cite{lu2024openkd} for comparison. Due to the unavailability of open-source code bases for these works, we implement them to the best of our ability and compare them in \cref{tab:supp_sota}. 

\begin{table}[t]
    \scriptsize
    \centering
    \setlength{\tabcolsep}{2.2pt}
    \begin{tabular}{lllcccccg}
    \toprule
    
    \multirow{2}{*}{Class}  & \multirow{2}{*}{Keypoints} & \multirow{2}{*}{Methods}     & \multicolumn{6}{c}{PCK@0.1} \\
    \cmidrule{4-9}
                            &                            &                                  & Cat   & Cow   & Dog   & Horse & Sheep & \textbf{Mean}  \\
    \midrule
    \multirow{8}{*}{Seen}   & \multirow{4}{*}{Base}      & GeometryKP~\cite{he2023few}      & 33.47 & 22.57 & 26.96 & 27.08 & 19.84 & 25.98          \\
                            &                            & ProbIntr~\cite{novotny2018self}  & 47.93 & 36.77 & 41.11 & 42.47 & 32.92 & 40.24          \\
                            &                            & OpenKD~\cite{lu2024openkd}       & 60.44 & 46.28 & 50.98 & 51.85 & 42.03 & 50.32          \\
                            &                            & \textbf{Proposed}                & \textbf{67.34} & \textbf{49.89} & \textbf{56.28} & \textbf{56.35} & \textbf{45.65} & \textbf{55.10} \\
    \cmidrule{2-9}
                            & \multirow{4}{*}{Novel}     & GeometryKP~\cite{he2023few}      & 22.07 & 14.18 & 19.65 & 12.71 & 17.39 & 17.20          \\
                            &                            & ProbIntr~\cite{novotny2018self}  & 31.29 & 24.53 & 28.37 & 22.83 & 26.33 & 26.67          \\
                            &                            & OpenKD~\cite{lu2024openkd}       & 49.92 & 38.27 & 42.74 & 38.46 & 34.17 & 40.71          \\
                            &                            & \textbf{Proposed}                & \textbf{55.69} & \textbf{43.09} & \textbf{46.58} & \textbf{43.94} & \textbf{36.39} & \textbf{45.14} \\
    \midrule
    \multirow{8}{*}{Unseen} & \multirow{4}{*}{Base}      & GeometryKP~\cite{he2023few}      & 27.39 & 23.76 & 19.51 & 26.93 & 20.58 & 23.63          \\
                            &                            & ProbIntr~\cite{novotny2018self}  & 40.76 & 36.13 & 30.15 & 42.84 & 34.52 & 36.88          \\
                            &                            & OpenKD~\cite{lu2024openkd}       & 44.96 & 41.54 & 36.49 & 43.18 & 39.06 & 41.05          \\
                            &                            & \textbf{Proposed}                & \textbf{47.36} & \textbf{42.97} & \textbf{38.30} & \textbf{46.17} & \textbf{41.03} & \textbf{43.17} \\
    \cmidrule{2-9}
                            & \multirow{4}{*}{Novel}     & GeometryKP~\cite{he2023few}      & 14.25 &  8.69 & 11.81 & 13.44 &  9.37 & 11.51          \\
                            &                            & ProbIntr~\cite{novotny2018self}  & 23.91 & 17.59 & 18.37 & 13.85 & 21.63 & 19.07          \\
                            &                            & OpenKD~\cite{lu2024openkd}       & 40.38 & 39.72 & 36.07 & 35.91 & 34.67 & 37.35          \\
                            &                            & \textbf{Proposed}                & \textbf{44.42} & \textbf{40.13} & \textbf{36.91} & \textbf{37.77} & \textbf{35.77} & \textbf{39.00} \\
    \bottomrule
    \end{tabular}
    \vspace{-1em}
    \caption{Additional quantitative comparison of the state-of-the-art strategies with the proposed framework in $K=1$ shot setting delineating the superiority of the proposed method in overall performance on Animal Pose~\cite{cao2019cross} dataset.
    \vspace{-1em}
    }
    \label{tab:supp_sota}
\end{table}

\end{document}